\documentclass{article}

\usepackage{microtype}
\usepackage{graphicx}
\usepackage{subcaption}
\usepackage{booktabs} % for professional tables

\usepackage{hyperref}

\usepackage[accepted]{icml2026}

\usepackage{amsmath}
\usepackage{amssymb}
\usepackage{mathtools}
\usepackage{amsthm}

\usepackage{adjustbox}
\usepackage{multirow}

\usepackage{amsmath,amssymb,amsfonts}
\usepackage{bm}

\usepackage[capitalize,noabbrev]{cleveref}

\theoremstyle{plain}

\theoremstyle{definition}

\theoremstyle{remark}

\usepackage[textsize=tiny]{todonotes}

\icmltitlerunning{Rethinking 3D Shape Generation: Diffusion over Superquadrics}

\begin{document}

\twocolumn[{
\renewcommand\twocolumn[1][]{#1}%
  \icmltitle{Rethinking 3D Shape Generation: Diffusion over Superquadrics}



  \icmlsetsymbol{equal}{*}

  \begin{icmlauthorlist}
    \icmlauthor{Zhiyang Liu}{arc,nus}
    \icmlauthor{Wanze Li}{nus}
    \icmlauthor{Yuwei Wu}{nus}
    \icmlauthor{Chengran Yuan}{arc,nus}
    \icmlauthor{Jiawei Sun}{arc,nus}
    \icmlauthor{Rui Zheng}{sutd}
    \icmlauthor{Marcelo H Ang Jr}{arc,nus}
  \end{icmlauthorlist}

  \icmlaffiliation{arc}{Advanced Robotics Centre, National University of Singapore}
  \icmlaffiliation{nus}{College of Design and Engineering, National University of Singapore}
  \icmlaffiliation{sutd}{Engineering Systems and Design, Singapore University of Technology and Design}

  \icmlcorrespondingauthor{Zhiyang Liu}{iszhiyang@gmail.com}

  \icmlkeywords{Machine Learning, ICML}

  \vskip 0.3in

% \begin{center}
%   \includegraphics[width=1\textwidth]{figure/teaser.pdf}
%   \captionof{figure}{Enter Caption}
%   \label{fig:placeholder}
% \end{center}

}]



\printAffiliationsAndNotice{}  % no special notice (required even if empty)

\begin{abstract}
\noindent

Diffusion models have advanced 3D shape generation, yet most methods still denoise in high-cardinality spaces (e.g., voxel/SDF grids, meshes, or point clouds), which is computationally and memory intensive and makes it difficult to scale in terms of both higher resolution and stronger controllability.
We rethink the diffusion representation and propose to move diffusion from dense geometry to compact geometric primitives, representing each shape as a small set of \textbf{superquadrics}.
Instead of operating on thousands to millions of geometric representation values, we leverage 7KB superquadric parameters (pose, size, and shape), drastically reducing diffusion-state dimensionality and per-step compute/memory.
Our diffusion-over-superquadrics improves scalability by supporting broader capabilities (e.g., resolution-free point-cloud decoding, part-level editing, and constraint-based design) and achieving competitive surface-fidelity and distributional performance on standard benchmarks after point-cloud decoding, while enabling efficient generation within 0.6s per shape for most conditions.

\end{abstract}

\begin{figure*}
  \includegraphics[width=1\textwidth]{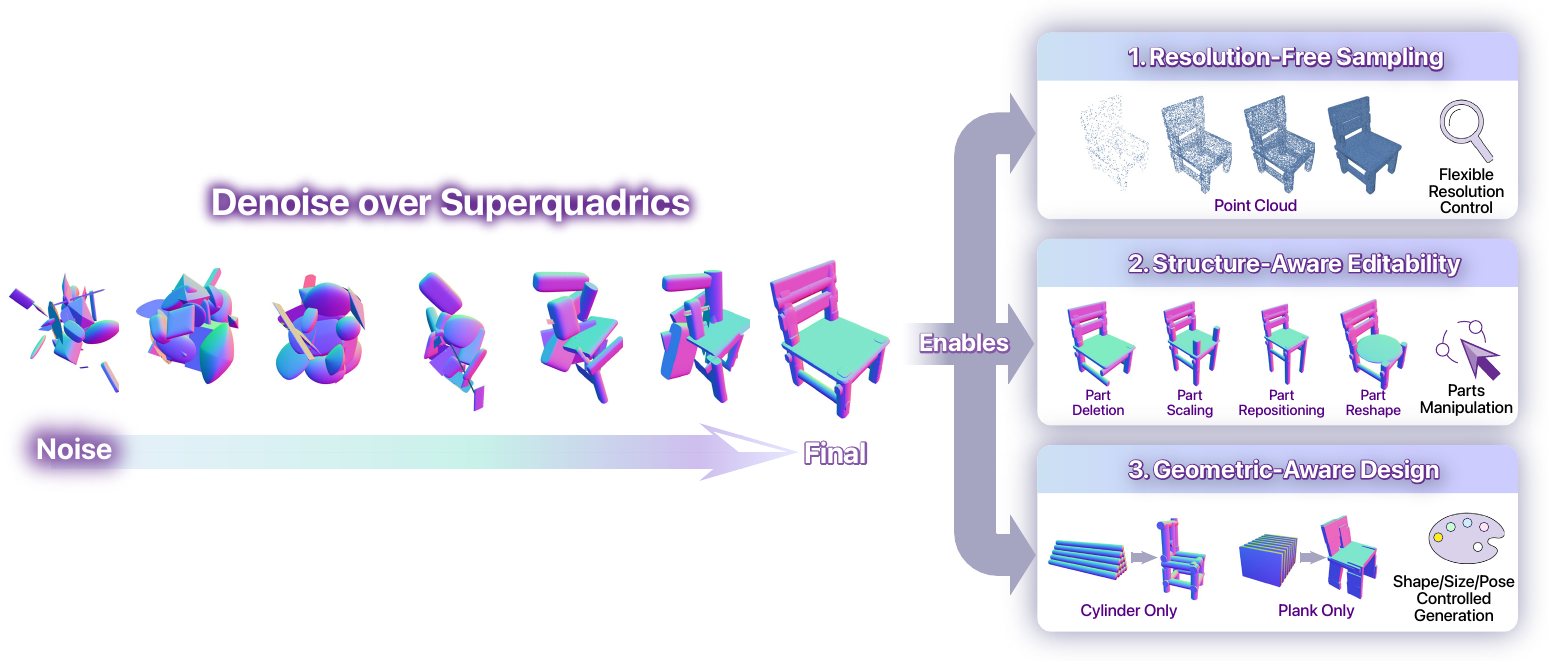}
  \captionof{figure}{Denoise over superquadrics. Starting from noise, our model denoises a compact set of superquadric primitives to a final structured shape, which can be decoded into point clouds at arbitrary resolution. The explicit primitive parameters further enable structure-aware editability (part deletion, scaling, repositioning, and reshaping) and geometric-aware design via constrained denoising process (e.g., cylinder-only or plank-only generation).}
  \label{fig:teaser}
  \vspace{-2mm}
\end{figure*}

\section{Introduction}
Diffusion models generate samples by learning the denoising process~\cite{ho2020ddpm,song2020score}, and recent Transformer-based designs have improved scalability to larger models and datasets~\cite{peebles2023scalable,bao2023one}. This progress has enabled 3D diffusion for shape generation, from unconditional to image-to-3D settings~\cite{wang2025diffusion}. Yet a central bottleneck remains: \emph{the representation that diffusion is asked to model.}
% These advances have spurred 3D diffusion methods for shape generation, from unconditional to image-to-3D~\cite{wang2025diffusion}. However, as highlighted in recent surveys~\cite{wang2025diffusion}, 3D diffusion still faces a persistent bottleneck: \emph{the representation we ask diffusion to model.}

Most 3D diffusion methods denoise high-cardinality representations. Point-based models denoise thousands of continuous values per shape (e.g., 2{,}048-point clouds are standard in ShapeNet benchmarks)~\cite{zeng2022lion,chang2015shapenet}, while grid-based discretizations can be far larger: sparse voxel generators explicitly target millions of voxels~\cite{ren2024xcube}, and tetrahedral partitions scale cubically with resolution (e.g., $0.72\!\cdot\!R^3$ elements, evaluated up to $R{=}192$)~\cite{kalischek2024tetradiffusion}. 
Because denoising is repeated for tens to hundreds of steps, compute, memory, and runtime scale sharply with the dimensionality of the diffusion state.
Moreover, when the state itself is high-cardinality, model capacity is often spent on dense geometric bookkeeping rather than object-level structure.
In response, recent lines of work improve efficiency by changing the diffusion space, e.g., via hierarchical discretizations such as octrees~\cite{xiong2025octfusion} or compact latent parameterizations such as tri-planes~\cite{wu2024direct3d}. However, these approaches expose a familiar trade-off: discretization-based methods remain resolution-dependent, while latent diffusion often sacrifices explicit geometric interpretability and direct editability for compactness. 
This leads to a natural question:
\begin{quote}
Can we scale diffusion-based 3D generation while retaining an explicit and compact representation?
\end{quote}

% We answer this question by rethinking the diffusion representation. Rather than denoising dense geometric fields, we perform \textbf{diffusion over superquadrics} (\textbf{DoSs}), representing each shape as a set of geometric primitives with continuous parameters for shape ($\mathbb{R}^2$), size ($\mathbb{R}^3$), and pose ($\mathrm{SE}(3)$).
% This primitive parameter representation is explicit. Each token corresponds to an interpretable geometric component. It is also compact because DoSs denoises a fixed size token representation per shape, instead of denoising geometry at a chosen voxel resolution or point count.
% Recent work has revisited superquadrics as practical building blocks for part structured 3D representations and structured reasoning at scale~\cite{fedele2025superdec,zuo2025quadricformer}. This motivates using superquadrics as generative tokens in DoSs.
% 3D generation then becomes denoising a short token sequence, encouraging the model to capture part structure and inter part relationships directly.

We answer this question by rethinking the diffusion representation. Rather than denoising dense geometric fields, we perform \textbf{diffusion over superquadrics} (\textbf{DoSs}), representing each shape as a set of geometric primitives with continuous parameters for shape ($\mathbb{R}^2$), size ($\mathbb{R}^3$), and pose ($\mathrm{SE}(3)$).
This primitive parameter representation is explicit. Each token corresponds to an interpretable geometric component. It is also compact because DoSs denoises a fixed size token representation per shape, instead of denoising geometry at a chosen voxel resolution or point count.
Recent work has revisited superquadrics as practical building blocks for part structured 3D representations and structured reasoning at scale~\cite{fedele2025superdec,zuo2025quadricformer}. This motivates using superquadrics as generative tokens in DoSs.
3D generation then becomes denoising a short token sequence, encouraging the model to capture part structure and inter part relationships directly.

Based on this perspective, we introduce a diffusion model that generates superquadric parameters directly, but a naive implementation is not straightforward for three reasons. First, shapes require different numbers of primitives depending on geometric complexity, so we represent each shape by up to $K_{\max}$ primitives and include an additional existence score per primitive to handle variable primitive counts. Second, diffusion perturbs variables with Gaussian noise in an ambient Euclidean space~\cite{ho2020ddpm,song2020score}, but pose includes rotation on $\mathrm{SO}(3)$, which is a curved manifold and can be discontinuous under common parameterizations, so we follow~\citet{zhou2019continuity} and use a continuous 6D rotation representation mapped to a valid rotation matrix via orthonormalization. Third, our diffusion model~\cite{ho2020ddpm} uses a 1D convolutional U-Net backbone that operates on a fixed length ordered token tensor, so we impose a deterministic canonical ordering of primitives, by volume or position, to convert an unordered primitive set into a consistent token sequence for learning. Fig.~\ref{fig:teaser} overviews DoSs. Starting from Gaussian noise, our model denoises a compact set of superquadric primitives into a structured shape, which is then decoded into point clouds at arbitrary sampling resolution. The explicit primitive parameters also support part level edits and constrained denoising for design. 
Our contributions are threefold. 

% Our contributions are threefold.
% \begin{itemize}
%     \item \textbf{DoSs.} We introduce diffusion over superquadrics (DoSs), a 3D shape generator that denoises a compact and explicit set of superquadric primitives rather than dense voxel, mesh, or point representations.
%    \item \textbf{Making superquadrics diffusable.} We develop a stable diffusion formulation over superquadric primitives that handles variable primitive counts and $\mathrm{SO}(3)$ rotations, using an existence score, 6D rotations~\cite{zhou2019continuity}, and canonical ordering.
%     \item \textbf{Efficiency, quality, and scalability.} We show that DoSs is competitive on ShapeNet~\cite{chang2015shapenet} after point cloud decoding, while improving efficiency through a much smaller diffusion state, and we demonstrate scalable capabilities including resolution free decoding, part level edits, and constraint based design via constrained denoising.
% \end{itemize}
\vspace{-2mm}
\begin{itemize} 
    \item \textbf{Rethinking shape representation for diffusion.} We rethink 3D diffusion by shifting the denoising target from high cardinality voxel, mesh, or point representations to a compact and explicit set of superquadric primitives, and propose diffusion over superquadrics (DoSs) for 3D shape generation.\vspace{-1mm}

    \item \textbf{Making superquadrics diffusable.} We develop a practical diffusion formulation over superquadric primitives by addressing three implementation obstacles: variable primitive counts are handled with an existence score, rotations are diffused with a continuous 6D representation~\cite{zhou2019continuity}, and unordered primitives are converted into a consistent token sequence using deterministic canonical ordering.\vspace{-1mm}

    \item \textbf{Efficiency, quality, and scalable capabilities.} We show that DoSs achieves competitive fidelity and distributional performance on ShapeNet~\cite{chang2015shapenet}, while improving efficiency through a much smaller diffusion state. We further demonstrate capabilities enabled by DoSs, including resolution-free decoding, part level edits, and constraint based design via constrained denoising.\vspace{-1mm}
\end{itemize}

\section{Related Work}

\textbf{Diffusion for 3D shape generation.}
Recent 3D diffusion methods differ largely by the representation they denoise. A line of work performs diffusion over explicit 3D samples or discretizations such as point sets and voxelized tokens. Point E generates point clouds via cascaded diffusion with image conditioning~\citep{nichol2022point}, and DiT 3D shows that diffusion transformers can denoise voxelized point cloud tokens effectively~\citep{mo2023dit}. These explicit approaches provide straightforward geometry extraction and evaluation, but their generation cost often grows with sampling density or grid resolution, and enforcing part level edits or hard structural constraints typically requires additional design of conditioning mechanisms.

Another line of work improves scalability by moving diffusion to compact latent or implicit spaces and decoding to surfaces. Sin3DM and Direct3D use triplane latent representations and perform diffusion in that latent space~\citep{wu2023sin3dm,wu2024direct3d}. While such methods can decouple diffusion from dense 3D grids, the intermediate variables are often not directly interpretable, which makes explicit geometric control and constraint enforcement less direct.

\textbf{Superquadrics and primitive representations.}
Superquadrics are a classical family of parametric primitives that compactly represent a range of quadric like shapes~\citep{barr1981superquadrics}. Recent work revisits superquadrics as interpretable parts for reconstruction and decomposition. SuperDec decomposes objects and scenes into superquadric primitives from point clouds~\citep{fedele2025superdec}, ISCO constructs superquadric based object models from multiple views~\citep{alaniz2023iterative}, and Differentiable Blocks World represents scenes as textured superquadric meshes optimized through differentiable rendering~\citep{monnier2023differentiable}. These methods focus on reconstruction, fitting, or decomposition rather than learning a category level generative prior. In contrast, we perform diffusion directly over a set of superquadric primitives, yielding a compact and explicit denoising space that supports resolution free decoding and direct parameter level edits and constraints.

\section{Methodology}
\label{sec:method}

We represent each object as a set of superquadric primitives and learn a denoising diffusion model directly in this compact parameter space. The methodology has four parts: (i) revisiting superquadrics, (ii) revisiting Diffusion Probabilistic Models (DDPM), (iii) what makes superquadrics diffusable in practice, and (iv) how to scale diffusion over superquadrics to broader capabilities.

% ------------------------------------------------------------
\subsection{Revisiting Superquadrics}
\label{sec:method:sq}

\begin{figure}
  \includegraphics[width=0.45\textwidth]{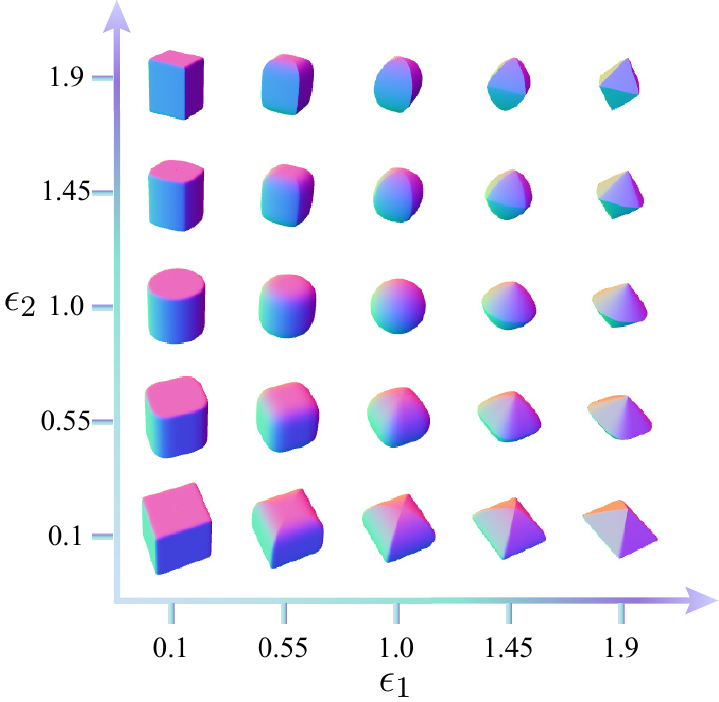}
  \captionof{figure}{A grid of superquadrics generated by varying the two shape exponents $(\epsilon_1,\epsilon_2)\in\{0.1,0.55,1.0,1.45,1.9\}^2$ (axes), while keeping pose and scale fixed.}
  \label{fig:primitives}
  \vspace{-2mm}
\end{figure}

% \paragraph{Superquadric as geometric primitives.}
% % We use the \emph{superellipsoid} (often referred to as a superquadric in vision/robotics) \citep{barr1981superquadrics}.
% Superquadrics are a class of geometric primitives with highly expressive shape representation capabilities, enabling the modeling of a wide variety of complex geometries.
% Let $\mathbf{x}=(x,y,z)^\top \in \mathbb{R}^3$ be a point in the local superquadrics frame. With axis lengths $\mathbf{a}=(a_1,a_2,a_3)^\top$ and shape defined by $\boldsymbol{\epsilon}=(\epsilon_1,\epsilon_2)^\top$,  (where $a_i>0$ and $\epsilon_j>0$), the inside--outside function is
% \begin{equation}
% \label{eq:sq-implicit-local}
% F(\mathbf{x};\mathbf{a},\boldsymbol{\epsilon})
% =
% \left(
% \left|\frac{x}{a_1}\right|^{\frac{2}{\epsilon_2}}
% +
% \left|\frac{y}{a_2}\right|^{\frac{2}{\epsilon_2}}
% \right)^{\frac{\epsilon_2}{\epsilon_1}}
% +
% \left|\frac{z}{a_3}\right|^{\frac{2}{\epsilon_1}}.
% \end{equation}
% The surface is $\{\mathbf{x}\mid F(\mathbf{x};\mathbf{a},\boldsymbol{\epsilon})=1\}$; points satisfying $F<1$ are inside the surface and $F>1$ are outside the surface.

\textbf{Superquadrics as geometric primitives.}
Superquadrics are a family of compact geometric primitives that  interpolate smoothly between round and boxy shapes, making them a flexible continuous representation for 3D geometry. Fig.~\ref{fig:primitives} illustrates this effect by varying the two shape exponents $(\epsilon_1,\epsilon_2)$ while keeping pose and scale fixed.
Let $\mathbf{x}=(x,y,z)^\top\in\mathbb{R}^3$ denote a point in the local coordinate frame of a superquadric. A superquadric is parameterized by axis lengths $\mathbf{a}=(a_1,a_2,a_3)^\top$ and shape exponents $\boldsymbol{\epsilon}=(\epsilon_1,\epsilon_2)^\top$, where $a_i>0$ and $\epsilon_j>0$. Its inside-outside function is shown in Eq.~\ref{eq:sq-implicit-local}. The surface is given by $\{\mathbf{x}\mid F(\mathbf{x};\mathbf{a},\boldsymbol{\epsilon})=1\}$. Points with $F(\mathbf{x};\mathbf{a},\boldsymbol{\epsilon})<1$ lie inside the shape, and those with $F(\mathbf{x};\mathbf{a},\boldsymbol{\epsilon})>1$ lie outside.
\begin{equation} \vspace{-1mm}
\label{eq:sq-implicit-local}
F(\mathbf{x};\mathbf{a},\boldsymbol{\epsilon})
=
\left(
\left|\frac{x}{a_1}\right|^{\frac{2}{\epsilon_2}}
+
\left|\frac{y}{a_2}\right|^{\frac{2}{\epsilon_2}}
\right)^{\frac{\epsilon_2}{\epsilon_1}}
+
\left|\frac{z}{a_3}\right|^{\frac{2}{\epsilon_1}}.
\end{equation} \vspace{-1mm}

\textbf{Pose in $SE(3)$.}
Each primitive has a rigid pose $(\mathbf{R},\mathbf{t})$ with $\mathbf{R}\in \mathrm{SO}(3)$ and $\mathbf{t}\in\mathbb{R}^3$.
For a point $\mathbf{p}\in\mathbb{R}^3$ in the world frame, local coordinates are
$\label{eq:sq-world-to-local}
\mathbf{x} = \mathbf{R}^\top(\mathbf{p}-\mathbf{t}),$ and the posed implicit surface is $F(\mathbf{R}^\top(\mathbf{p}-\mathbf{t});\mathbf{a},\boldsymbol{\epsilon})=1$.

\textbf{Superquadrics volume (for canonical ranking).}
We use a closed-form volume for superquadrics~\citep{jaklic2000segmentation}:
\begin{equation}
\label{eq:sq-volume}
\mathrm{Vol}(\mathbf{a},\boldsymbol{\epsilon})
=
2 a_1 a_2 a_3 \, \epsilon_1 \epsilon_2\;
\mathrm{B}(\frac{\epsilon_1}{2},\epsilon_1+1)\,
\mathrm{B}(\frac{\epsilon_2}{2},\frac{\epsilon_2+2}{2}).
\end{equation}
Here $\mathrm{B}(p,q)=\Gamma(p)\Gamma(q)/\Gamma(p+q)$, where the Gamma function is defined for $t>0$ as $\Gamma(t)=\int_{0}^{\infty} u^{t-1}e^{-u}\,du$.

% ------------------------------------------------------------
\subsection{Revisiting DDPM}
\label{sec:method:ddpm}

In this work, we adopt DDPM \citep{sohl2015deep,ho2020ddpm} to generate new data $\mathbf{x}_0$ that obey the distribution of the training set. Concretely, DDPM recovers the output $\mathbf{x}_0$ from a Guassian noise $\mathbf{x}_T$ by iteratively denoising with a noise prediction network $\boldsymbol{\xi}_\theta$
\begin{equation}
\label{eq:ddpm-reverse}
\mathbf{x}_{t-1}
=
\gamma_t(\mathbf{x}_{t} - \beta_t\boldsymbol{\xi}_\theta(\mathbf{x}_t, t))+z_t,
\end{equation}
Here $t = T, T-1, ..., 1$ is the index for the iteration. $z_t$ is a Gaussian noise with zero mean and $\gamma_t, \beta_t$ are parameters determined by timestep. 

The DDPM is trained in the forward (noising) process, where the Gaussian noise determined by $t$ is added to the training data iteratively $q(\mathbf{x}_t\mid \mathbf{x}_{t-1})
=
\mathcal{N}\!\left(\sqrt{\alpha_t}\mathbf{x}_{t-1},\, (1-\alpha_t)\mathbf{I}\right)$, which implies 
\begin{equation}
\label{eq:ddpm-eps}
\mathbf{x}_t = \sqrt{\bar{\alpha}_t}\mathbf{x}_0 + \sqrt{1-\bar{\alpha}_t}\,\boldsymbol{\xi}.
\end{equation}
where $\boldsymbol{\xi}\sim\mathcal{N}(\mathbf{0},\mathbf{I})$, $\bar{\alpha}_t=\prod_{s=1}^{t}\alpha_s$, and $\alpha_t$ is linearly determined by $t$ in this work. 
The noise prediction network $\boldsymbol{\xi}_\theta$ is trained to learn the noise term in Eq.\ref{eq:ddpm-eps} and the simplified training objective is
\begin{equation}
\label{eq:ddpm-loss}
\mathcal{L}_{\mathrm{DDPM}}
=
\mathbb{E}_{t,\mathbf{x}_0,\boldsymbol{\xi}}
\left[\left\|\boldsymbol{\xi}-\boldsymbol{\xi}_\theta(\mathbf{x}_t,t)\right\|_2^2\right].
\end{equation}
As proven in \cite{ho2020ddpm}, minimizing Eq.\ref{eq:ddpm-loss} is equivalent to minimizing the KL-divergence between the actual distribution of the training set and the distribution of data generated using Eq.\ref{eq:ddpm-reverse}.

\subsection{What Makes Superquadrics Diffusable?}
\label{sec:method:diffusable}

This section presents three design choices that make diffusion over superquadrics well defined and practical: (i) existence scores to handle variable primitive counts, (ii) a continuous 6D rotation parameterization for pose diffusion, and (iii) deterministic canonical ordering to obtain a consistent token sequence.

\subsubsection{Tokenization via existence scores}
\label{sec:method:set2tensor}

\textbf{Variable primitive counts.}
Different objects are decomposed into different numbers of superquadric primitives.
Let object $i$ contain $K_i$ primitives, where $K_i$ varies across the dataset.
Since DDPM requires a fixed-dimensional state, we fix a global maximum $K_{\max}$ and represent every object with $K_{\max}$ slots.
To retain the original cardinality, each slot carries an existence score.

\textbf{Token with existence.}
For slot $k$, we store translation $\mathbf{t}_k\in\mathbb{R}^3$, a 6D rotation parameter $\mathbf{r}_k\in\mathbb{R}^6$ (decoded in Sec.~\ref{sec:method:se3}), axis length $\mathbf{a}_k\in\mathbb{R}^3$, unconstrained shape variables $\boldsymbol{\epsilon}_k\in\mathbb{R}^2$, and an existence score $e_k\in\mathbb{R}$:
\begin{equation}
\label{eq:token_existence}
\tilde{\mathbf{z}}_k
=
\big[\ \mathbf{t}_k^\top,\ \mathbf{r}_k^\top,\ \mathbf{a}_k^\top,\ \boldsymbol{\epsilon}_k^\top,\ e_k\ \big]^\top
\in \mathbb{R}^{15}.
\end{equation}

\textbf{Fixed-size state.}
We pad each sample to $K_{\max}$ rows using all-zero rows. For each slot, we set $e_k=1$ if it corresponds to a valid primitive and $e_k=0$ otherwise, producing a fixed-size matrix $\tilde{\mathbf{Z}}_0\in\mathbb{R}^{K_{\max}\times 15}$.
This yields a Euclidean diffusion state while preserving object-specific cardinality through the existence channel. In Sec.~\ref{sec:ablation}, we ablate this design against a copy-paste padding baseline that fills missing slots by duplicating existing primitives to reach $K_{\max}$.

\subsubsection{$\mathrm{SO}(3)$ diffusion via 6D rotations}
\label{sec:method:se3}

\textbf{Continuity and ambiguity.}
DDPM assumes the state evolves smoothly in Euclidean coordinates, but common rotation parameterizations are not globally well-behaved.
Euler angles satisfy periodic identifications, e.g.,
$\label{eq:euler_discontinuity}
\mathbf{R}(\boldsymbol{\theta}) = \mathbf{R}(\boldsymbol{\theta}+2\pi\mathbf{e}_j),$
and unit quaternions exhibit the antipodal symmetry $q\sim -q$.
Such non-unique / discontinuous coordinates can cause the forward noising process to mix distinct rotations and make denoising unstable.

\textbf{6D rotations.}
Primitive orientations are stored as ZYX Euler angles $\boldsymbol{\phi}\in\mathbb{R}^3$. Since diffusion perturbs variables with Gaussian noise in an ambient Euclidean space, we avoid diffusing Euler angles directly and instead diffuse the continuous 6D rotation representation of~\citet{zhou2019continuity}. We first convert Euler angles to a rotation matrix
$\mathbf{R}=\mathcal{R}_{\mathrm{ZYX}}(\boldsymbol{\phi})\in\mathrm{SO}(3)$, then map $\mathbf{R}$ to a 6D vector by stacking its first two columns:
$\mathbf{a}_1=\mathbf{R}_{:,1}$, $\mathbf{a}_2=\mathbf{R}_{:,2}$, and $\mathbf{r}=[\mathbf{a}_1;\mathbf{a}_2]\in\mathbb{R}^6$.
During training and sampling, diffusion is applied to $\mathbf{r}$ in $\mathbb{R}^6$.
When forming poses or evaluating geometry, we decode $\mathbf{r}$ back to a valid rotation matrix by Gram Schmidt orthonormalization. Specifically, we normalize $\mathbf{a}_1$ to obtain $\mathbf{b}_1=\mathrm{norm}(\mathbf{a}_1)$, remove the component of $\mathbf{a}_2$ along $\mathbf{b}_1$ and normalize to obtain
$\mathbf{b}_2=\mathrm{norm}(\mathbf{a}_2-(\mathbf{b}_1^\top\mathbf{a}_2)\mathbf{b}_1)$, then set $\mathbf{b}_3=\mathbf{b}_1\times\mathbf{b}_2$.
The decoded rotation matrix is
\begin{equation}
\label{eq:rot6d}
\mathbf{R}(\mathbf{r})=[\mathbf{b}_1\ \mathbf{b}_2\ \mathbf{b}_3]\in\mathrm{SO}(3).
\end{equation}
This construction guarantees $\mathbf{R}(\mathbf{r})$ is orthonormal with determinant $1$, while keeping the diffusion variable $\mathbf{r}$ in a continuous Euclidean space. We ablate Euler angle diffusion versus the 6D representation in Sec.~\ref{sec:ablation}.

% \paragraph{6D representation.}
% We adopt the continuous 6D rotation representation of \citet{zhou2019continuity}.
% Given $\mathbf{r}=[\mathbf{a}_1;\mathbf{a}_2]\in\mathbb{R}^6$ with $\mathbf{a}_1,\mathbf{a}_2\in\mathbb{R}^3$, we construct
% \begin{equation}
% \begin{aligned}
% \label{eq:rot6d}
% \mathbf{b}_1 &= \frac{\mathbf{a}_1}{\|\mathbf{a}_1\|_2},\\
% \tilde{\mathbf{b}}_2 &= \mathbf{a}_2 - (\mathbf{b}_1^\top \mathbf{a}_2)\mathbf{b}_1,\qquad
% \mathbf{b}_2 = \frac{\tilde{\mathbf{b}}_2}{\|\tilde{\mathbf{b}}_2\|_2},\\
% \mathbf{b}_3 &= \mathbf{b}_1 \times \mathbf{b}_2,\qquad
% \mathbf{R}(\mathbf{r}) = [\mathbf{b}_1\ \mathbf{b}_2\ \mathbf{b}_3]\in \mathrm{SO}(3).
% \end{aligned}
% \end{equation}
% Diffusion is performed over the Euclidean variable $\mathbf{r}\in\mathbb{R}^6$, and geometric computations use the decoded $\mathbf{R}(\mathbf{r})$.

\subsubsection{Suppressing permutation}
\label{sec:method:canonical}

% \paragraph{Permutation ambiguity.}
% Even after fixing size, the superquadric representation of an object is fundamentally a \emph{set}: permuting primitive indices does not change the shape.
% However, DDPM is trained on an ordered tensor $\tilde{\mathbf{Z}}_0$.
% If the same object appears with arbitrary permutations, the learning target becomes unnecessarily multi-modal:
% \begin{equation}
% \label{eq:perm_ambiguity}
% \{\tilde{\mathbf{z}}_1,\ldots,\tilde{\mathbf{z}}_{K}\}
% \equiv
% \{\tilde{\mathbf{z}}_{\pi(1)},\ldots,\tilde{\mathbf{z}}_{\pi(K)}\},
% \quad \forall\ \pi\in\mathcal{S}_K,
% \end{equation}
% where $\mathcal{S}_K$ denotes the set of all permutations of $K$ elements. Such property injects permutation noise into denoising and can cause token identity drift across samples.

\textbf{Permutation ambiguity.}
Even after fixing the tensor size, the superquadric representation of an object is fundamentally a set: permuting primitive indices does not change the represented shape.
However, our diffusion model~\cite{ho2020ddpm} operates on an ordered token tensor $\tilde{\mathbf{Z}}_0$.
If the same object appears with arbitrary permutations, the data distribution in token space contains multiple equally valid orderings, which can hinder learning:
$\label{eq:perm_ambiguity}
\{\tilde{\mathbf{z}}_1,\ldots,\tilde{\mathbf{z}}_{K}\}
\equiv
\{\tilde{\mathbf{z}}_{\pi(1)},\ldots,\tilde{\mathbf{z}}_{\pi(K)}\},
\quad \forall\ \pi\in\mathcal{S}_K,$
where $\mathcal{S}_K$ is the set of all permutations of $K$ elements. This permutation ambiguity injects ordering noise into denoising and can lead to inconsistent token identities across samples.

% \paragraph{Deterministic canonicalization.}
% We impose a deterministic ordering of primitives \emph{before} padding so that each shape maps to a consistent tensor representation.
% In practice, we consider two simple canonicalization strategies and report an ablation between them in Sec.~\ref{sec:ablation}:
% \begin{itemize}
% \item \textbf{Volume-based.} Compute primitive volumes $V_k=\mathrm{Vol}(\mathbf{a}_k,\boldsymbol{\epsilon}_k)$ via Eq.~\eqref{eq:sq-volume}, then sort tokens by decreasing $V_k$.
% \item \textbf{Position-based (+Y rank).} We follow ShapeNet's canonical orientation convention \cite{chang2015shapenet} ($+Y$ up, $-Z$ front) and rank primitives by $+Y$ up. Let $\mathbf{t}_k=(t_{k,1},t_{k,2},t_{k,3})^\top$ be the primitive center and $\mathbf{e}_y=(0,1,0)^\top$. Define
% \begin{equation}
% \label{eq:y-rank}
% s_k=\mathbf{e}_y^\top \mathbf{t}_k=t_{k,2},
% \end{equation}
% and sort by increasing $s_k$ (bottom-to-top).
% \end{itemize}
% Either choice removes permutation-induced variance and yields consistent supervision, while the denoiser must still learn non-trivial inter-primitive relations (co-occurrence and relative pose) to produce coherent shapes.
\textbf{Deterministic canonicalization.}
We impose a deterministic primitive ordering before padding so each shape maps to a consistent token tensor. We consider two strategies and ablate them in Sec.~\ref{sec:ablation}. (i) Volume based: compute $V_k=\mathrm{Vol}(\mathbf{a}_k,\boldsymbol{\epsilon}_k)$ using Eq.~\ref{eq:sq-volume} and sort tokens by decreasing $V_k$. (ii) Position based: rank primitives by height using the $Y$ coordinate of their centers, $s_k=\mathbf{e}_y^\top\mathbf{t}_k=t_{k,2}$ with $\mathbf{e}_y=(0,1,0)^\top$, and sort by increasing $s_k$. Both reduce permutation variance and provide consistent supervision.

% ------------------------------------------------------------
\subsection{How to Scale Diffusion-over-Superquadrics?}
\label{sec:method:scale}

Our superquadrics diffusion model outputs a compact primitive set in token form (Eq.~\ref{eq:token_existence}), which enables several \emph{diffusion-based generation tasks} beyond unconditional sampling.
We focus on three capabilities that ``scale'' the generator in practice: (i) resolution-free surface sampling to point clouds, (ii) structure-aware editability, and (iii) geometric-aware design via constrained denoising. Examples about these capabilities are displayed in Fig.\ref{fig:teaser}.

\subsubsection{Resolution-free sampling}
\label{sec:method:resfree}

Given a generated tensor $\tilde{\mathbf{Z}}_0\in\mathbb{R}^{K_{\max}\times 15}$, we first select valid primitives by thresholding the existence score:
\begin{equation}
\label{eq:valid_slots}
\mathcal{K}=\{\,k\in\{1,\dots,K_{\max}\}: e_k>\tau_e\,\}.
\end{equation}
For each $k\in\mathcal{K}$, we decode the rotation matrix $\mathbf{R}_k=\mathbf{R}(\mathbf{r}_k)$ using Eq.~\eqref{eq:rot6d}.

\textbf{Per-primitive surface sampling.}
We sample each primitive surface using the standard superquadric parameterization with the signed power operator
$\operatorname{spow}(x,\epsilon)=\operatorname{sign}(x)\,|x|^{\epsilon}$.
For angles $(\eta,\omega)\in[-\frac{\pi}{2},\frac{\pi}{2}]\times[-\pi,\pi]$, the local surface point and its world coordinate are
\[
\mathbf{x}_k(\eta,\omega)=
\begin{bmatrix}
a_{k1}\ \operatorname{spow}(\cos\eta,\epsilon_{k1})\ \operatorname{spow}(\cos\omega,\epsilon_{k2})\\
a_{k2}\ \operatorname{spow}(\cos\eta,\epsilon_{k1})\ \operatorname{spow}(\sin\omega,\epsilon_{k2})\\
a_{k3}\ \operatorname{spow}(\sin\eta,\epsilon_{k1})
\end{bmatrix},
\]
$\mathbf{p}_k(\eta,\omega)=\mathbf{R}_k\,\mathbf{x}_k(\eta,\omega)+\mathbf{t}_k$.

Following the equal distance and arc length sampling procedure of \citet[][Suppl., Sec.~4]{liu2022robust}, we generate a candidate set
$\mathcal{P}_k=\{\mathbf{p}_k(\eta_n,\omega_n)\}_{n=1}^{M_k}$ for each primitive and merge
$\mathcal{P}=\bigcup_{k\in\mathcal{K}}\mathcal{P}_k$.

\textbf{Filtering points inside overlaps.}
Because primitives can overlap, $\mathcal{P}$ may contain points that lie inside other primitives and should not appear on the surface of the composed shape. For a candidate point $\mathbf{p}\in\mathcal{P}_k$, we keep it only if it is outside every other primitive:
$\label{eq:sq_overlap_filter}
F_j\!\left(\mathbf{R}_j^\top(\mathbf{p}-\mathbf{t}_j);\mathbf{a}_j,\boldsymbol{\epsilon}_j\right)\ge 1,
\quad \forall j\in\mathcal{K}\setminus\{k\},$
where $F_j(\cdot)$ is the inside outside function in Eq.~\eqref{eq:sq-implicit-local}. The remaining set is denoted by $\mathcal{P}^{\mathrm{surf}}$.

\textbf{Uniform density via FPS.}
Arc length sampling is only near uniform on a single superquadric. In our setting, non uniformity arises from two sources: (i) the arc length sampler itself is approximate, and (ii) merging surface samples from multiple primitives, especially near overlaps and intersections, creates locally dense regions. To produce an even density point cloud at any requested size $N$, we run farthest point sampling (FPS) on the filtered surface set $\mathcal{P}^{\mathrm{surf}}$. Starting from any $\mathbf{p}_1\in\mathcal{P}^{\mathrm{surf}}$, FPS iteratively adds $\mathbf{p}_n=\arg\max_{\mathbf{p}\in\mathcal{P}^{\mathrm{surf}}\setminus \mathcal{S}_{n-1}}\min_{\mathbf{q}\in\mathcal{S}_{n-1}}\|\mathbf{p}-\mathbf{q}\|_2$ and updates $\mathcal{S}_n=\mathcal{S}_{n-1}\cup\{\mathbf{p}_n\}$ until $|\mathcal{S}_N|=N$. This enables resolution free decoding, since the same generated superquadrics can be decoded into point clouds of arbitrary size without rerunning diffusion.

\subsubsection{Structure-aware editability}
\label{sec:method:edit}
Superquadrics expose explicit part-level parameters (pose, size, shape) and an existence channel, so edits can be applied directly in token space by modifying $\tilde{\mathbf{Z}}_0$. Typical operations include: (i) Part deletion: set $e_k\leftarrow \tau_e-\delta$ with $\delta>0$ so slot $k$ is removed by Eq.~\ref{eq:valid_slots}; (ii) Part scaling: set $\mathbf{a}_k\leftarrow \boldsymbol{\lambda}\odot\mathbf{a}_k$ with $\boldsymbol{\lambda}\in\mathbb{R}^3_{+}$; (iii) Part reposing: set $\mathbf{t}_k\leftarrow \mathbf{t}_k+\Delta\mathbf{t}$ and overwrite $\mathbf{r}_k$ with the 6D encoding of a desired rotation $\mathbf{R}'_k$ (Sec.~\ref{sec:method:se3}); (iv) Part reshaping: edit $\boldsymbol{\epsilon}_k$ to continuously control roundness and sharpness.

\subsubsection{Geometric-aware design}
\label{sec:method:design}

Beyond post-hoc edits, superquadric diffusion supports design-time control by constraining specific parameters throughout denoising.
Let $\Omega$ denote a set of geometric constraints expressed on token entries (or decoded parameters), and let $\Pi_{\Omega}$ be a projection operator that overwrites constrained channels with user-specified values.
During ancestral sampling, we apply
$\label{eq:constrained_denoise}
\tilde{\mathbf{Z}}_{t-1} \leftarrow \Pi_{\Omega} (\hat{\tilde{\mathbf{Z}}}_{t-1})$
for $t=T,T-1,\dots,1$, which is a simple instance of constraint-incorporating diffusion sampling \citep{chung2022diffusion}.

We take cylinder-only parts for furniture-like design as an example.
Suppose we want a subset of primitives $\mathcal{K}_{\mathrm{cyl}}\subseteq\mathcal{K}$ to behave like ``fixed-radius cylinders'' (i.e., circular cross-sections with prescribed roundness).
We can enforce
$\label{eq:cyl_constraints}
a_{k,1}=a_{k,2}= r_0, \ 
\boldsymbol{\epsilon}_k=\boldsymbol{\epsilon}^{\star},\
 \forall k\in\mathcal{K}_{\mathrm{cyl}},$
by clamping the corresponding entries.
More generally, $\Omega$ can specify part templates (e.g., ``all legs share the same radius''), symmetry constraints, or discrete part inclusion/exclusion through the existence channel.
Crucially, these controls operate on explicit, low-dimensional primitive parameters, making the design constraints interpretable and easy to enforce during diffusion.

\begin{table*}[t] \caption{
\textbf{Quantitative comparison on ShapeNet.}
We report 1-NNA@CD ($\downarrow$, ideal $\approx 50$) and COV@CD ($\uparrow$) for Chair, Airplane, and Car.
For each method, we list the diffusion state and its size (KB, float32), and inference time.
Baseline 1-NNA/COV values are from DiT-3D \cite{mo2023dit} and TIGER~\cite{ren2024tiger}.
Timing sources (RTX~3090, batch size 1): DPM/PVD from ~\cite{wu2023fast}, TIGER from~\cite{ren2024tiger}, LION from~\cite{zeng2022lion}, and DiT-3D is measured by us using the official checkpoint. Additional experimental results from multi-class training are provided in Appendix~\ref{multiclass}.
\textbf{Bold} and \underline{underline} denote best and second best.
}

\label{tab:quant_main} \centering \small \setlength{\tabcolsep}{3pt} \renewcommand{\arraystretch}{1.0} \begin{adjustbox}{max width=\textwidth} \begin{tabular}{l l c c cc cc cc} \toprule \multirow{2}{*}{Method} & \multirow{2}{*}{Diffusion state} & \multicolumn{1}{c}{Size} & \multicolumn{1}{c}{Inference} & \multicolumn{2}{c}{Chair} & \multicolumn{2}{c}{Airplane} & \multicolumn{2}{c}{Car} \\ & & \multicolumn{1}{c}{(KB)} & \multicolumn{1}{c}{(s/shape)} & \multicolumn{1}{c}{1-NNA} & \multicolumn{1}{c}{COV} & \multicolumn{1}{c}{1-NNA} & \multicolumn{1}{c}{COV} & \multicolumn{1}{c}{1-NNA} & \multicolumn{1}{c}{COV} \\ \cmidrule(lr){5-6}\cmidrule(lr){7-8}\cmidrule(lr){9-10} \midrule DPM~\cite{luo2021diffusion} & points ($2048{\times}3$) & \underline{24} & 22.8 & 60.05 & 44.86 & 76.42 & 48.64 & 68.89 & 44.03 \\ PVD~\cite{zhou20213d} & points ($2048{\times}3$) & \underline{24} & 29.9 & 57.09 & 36.68 & 73.82 & \underline{48.88} & 54.55 & 41.19 \\ TIGER~\cite{ren2024tiger} & points ($2048{\times}3$) & \underline{24} & \underline{9.73} & 54.61 & -- & 71.85 & -- & 54.31 & -- \\ LION~\cite{zeng2022lion} & latent ($128{+}8192$) & 33 & 27.12 & 53.70 & 48.94 & 67.41 & 47.16 & \underline{53.41} & \textbf{50.00} \\ MeshDiffusion~\cite{liu2023meshdiffusion} & hybrid ($64^3{\times}4$) & 4096 & -- & \underline{53.69} & 46.00 & 66.44 & 47.34 & 81.43 & 34.07 \\ DiT-3D~\cite{mo2023dit} & voxels ($32^3{\times}3$) & 384 & 12$^{*}$ & \textbf{49.11} & \textbf{52.45} & \underline{62.35} & \textbf{53.16} & \textbf{48.24} & \textbf{50.00} \\ \midrule Ours & superquadrics ($128{\times}15$) & \textbf{7} & \textbf{0.6} & 53.80 & \underline{51.31} & \textbf{61.92} & 46.77 & 57.50 & \underline{49.26} \\ \bottomrule \end{tabular} \end{adjustbox} \end{table*}

% \caption{ Quantitative comparison on ShapeNet. We report 1-NNA@CD ($\downarrow$, ideal $\approx 50$) and COV@CD ($\uparrow$). Unless otherwise noted, 1-NNA/COV values are taken from the CD columns of DiT-3D Table~1~\cite{mo2023dit}. TIGER reports 1-NNA@CD but not COV, thus we mark TIGER COV as n/r~\cite{ren2024tiger}. `Diffusion state'' indicates the denoised variable; Size (KB)'' assumes float32 and is computed from the diffusion state dimensionality (rounded). Inference time (s/shape) is reported when available: DPM/PVD are from the unified benchmark in Straight Flows (RTX~3090, batch size 1, Table~1)~\cite{wu2023fast}; TIGER time is from TIGER's timing table (average over categories)~\cite{ren2024tiger}; LION's 27.12~s is the 1000-step DDPM sampling time reported in LION~\cite{zeng2022lion}; DiT-3D time ($12^{*}$) is measured by us using the official DiT-3D implementation/checkpoint with batch size 1 and the default sampling setting. \textbf{Bold} indicates best and \underline{underline} indicates second best. }

% =========================================================
\section{Experiments}
\label{sec:experiments}

\subsection{Experimental Setup}
\label{sec:exp_setup}
\textbf{Dataset.} We follow the standard ShapeNet\cite{chang2015shapenet} used in prior work and report results on the \textsc{Chair}, \textsc{Airplane}, and \textsc{Car} categories.
Given a ShapeNet mesh, we extract a set of superquadric primitives for training and testing using Marching-Primitives~\cite{liu2023marching}, which fits superquadrics from a signed-distance representation.

\textbf{Evaluation metrics.}
We follow prior ShapeNet generation works~\cite{liu2023meshdiffusion,mo2023dit} and report distributional metrics 1-NNA and Coverage (COV) under Chamfer Distance (CD). Since our DoSs outputs a set of superquadric tokens, we first decode each generated shape into a point cloud and evaluate in point space: we sample 2{,}048 surface points per shape (matching the standard evaluation resolution) and compute CD between point clouds. \textbf{1-NNA} is the 1-nearest-neighbor classification accuracy for distinguishing generated samples from real samples under CD; values closer to $50\%$ are better, with $50\%$ indicating indistinguishability. \textbf{COV} measures generation diversity by the fraction of real shapes that are covered by the generated set under nearest neighbor matching, where a real shape is counted as covered if it is selected as the nearest neighbor of at least one generated shape; higher is better.

% \textbf{Evaluation Metrics.} We adopt the widely used distributional metrics \textbf{1-NNA} and \textbf{Coverage (COV)} under Chamfer Distance (CD), following prior ShapeNet generation evaluations~\cite{liu2023meshdiffusion,mo2023dit}.
% Since our diffusion space is \emph{superquadric tokens}, we first decode each generated token set into a point cloud for metric computation.
% Specifically, we sample \textbf{2048} surface points per generated shape (matching the standard evaluation resolution), and compute CD/EMD-based distances on these point clouds.

% \paragraph{1-NNA.}
% The 1-nearest-neighbor accuracy measures how well a classifier that assigns each sample the label of its nearest neighbor can distinguish generated samples from real samples; \emph{lower is better}, with $50\%$ indicating indistinguishability.

% \paragraph{Coverage (COV).}
% COV measures how many real shapes are ``covered'' by generated shapes within a nearest-neighbor matching under the chosen distance; \emph{higher is better}.

% \textbf{Implementation Details.} We train category-specific models for \textbf{4000 epochs} and finish training within approximately \textbf{2 hours per category} on a single GPU with \textbf{2\,GB} VRAM, highlighting the efficiency of diffusing in a compact superquadric token space.
% We use batch size \textbf{64}, optimizer \textbf{Adam}, and learning rate \textbf{$1\times10^{-4}$}. We sample with \textbf{100} denoising steps at inference time.

\textbf{Implementation details.}
We train category-specific models for 4{,}000 epochs using Adam with learning rate $1\times10^{-4}$ and batch size 64. Owing to the compact superquadric token diffusion space, training completes in about \textbf{2 hours per category} on a single RTX 4090 with \textbf{2\,GB VRAM usage}. At inference, we run 100 denoising steps and obtain a per-shape denoising time of \textbf{0.6\,s} with batch size 1. For each object, the $K_{\max}$ is set as 128 and $\tau_e = 0.5$.

% ---------------------------------------------------------
\begin{figure}
\centering
  \includegraphics[width=0.46\textwidth]{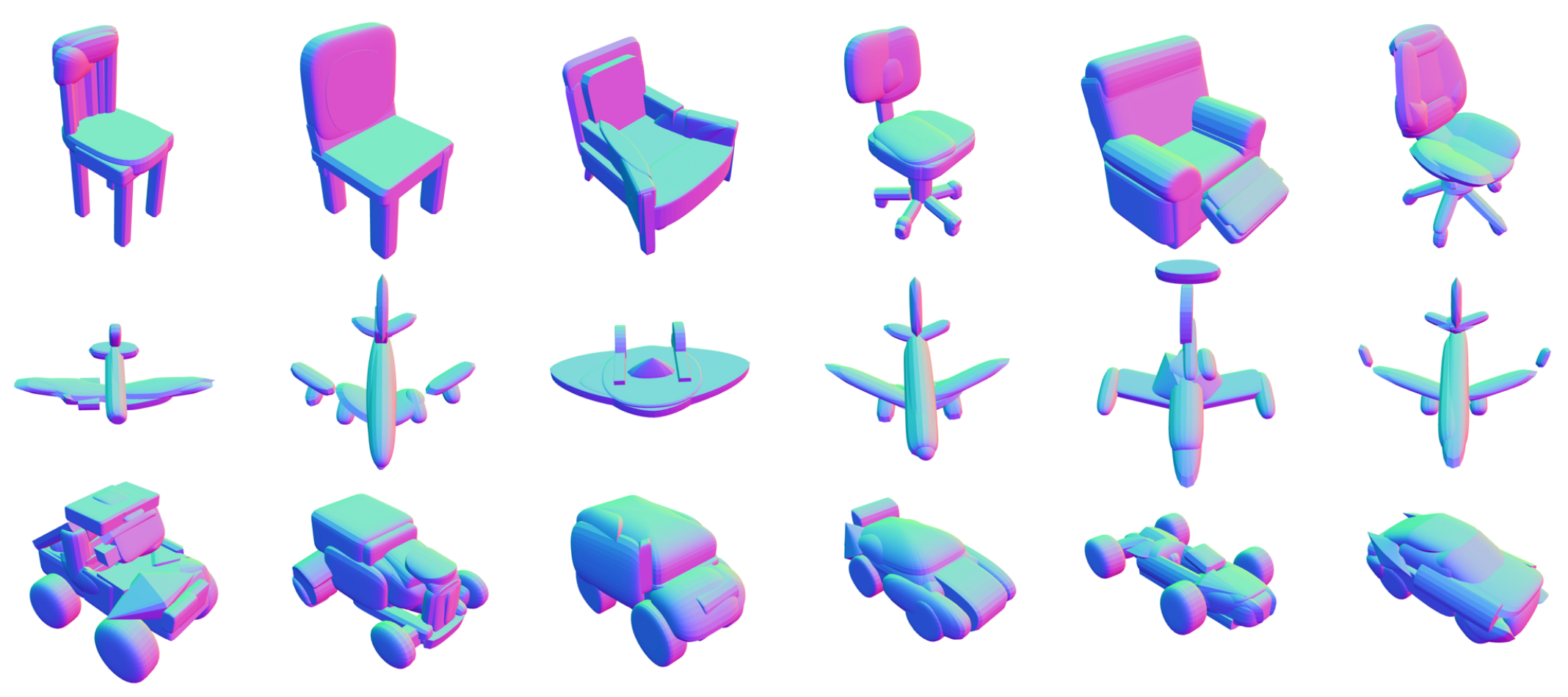}
  \captionof{figure}{Qualitative results from DoSs. Unconditional generations on ShapeNet categories (top to bottom: chair, airplane, car). More visualization for denoising and generation is shown in Appendix~\ref{appendix:qv}. }
  \label{fig:7}
  \vspace{-5mm}
\end{figure}

\subsection{Comparison with Existing Diffusion Models}
\label{sec:comparison}

Tab.~\ref{tab:quant_main} compares DoSs with representative diffusion baselines on ShapeNet \textsc{Chair}, \textsc{Airplane}, and \textsc{Car}, using standard distributional metrics 1-NNA@CD and COV@CD computed on 2{,}048-point clouds. Fig.~\ref{fig:7} shows unconditional samples generated by DoSs across the three categories.

% Tab.~\ref{tab:quant_main} compares DoSs to representative diffusion baselines on ShapeNet (\textsc{Chair}, \textsc{Airplane}, \textsc{Car}) using standard distributional metrics 1-NNA@CD and COV@CD computed on 2{,}048-point clouds. Fig.~\ref{fig:7} visualizes unconditional generations from DoSs for \textsc{Chair}, \textsc{Airplane}, and \textsc{Car}. We further evaluate a single multi-class DoSs model trained jointly on \textsc{Chair}, \textsc{Airplane}, and \textsc{Car} using category conditioning which is reproted in Appendix A. The joint model achieves similar performance as in Tab.~\ref{tab:quant_main} while it is trained in 6 hours on one RTX 4090. This suggests that DoSs is not restricted to category-specific training, although scaling to broader and more diverse datasets remains future work.
%For 1-NNA, values closer to $50\%$ are better (with $50\%$ indicating indistinguishability), while higher COV indicates better mode coverage and diversity.

\textbf{Comparison scope.} We emphasize that this comparison focuses on unconditional ShapeNet generation
under a matched point-cloud evaluation protocol. Several recent 3D generation
systems target different settings, such as text-to-3D or image-to-3D, and differ
in conditioning signals, training scale, output modality, and evaluation protocol.
Direct numerical comparison with these systems would therefore conflate the
effect of representation choice with conditioning strength and data scale.
Accordingly, our goal is not to claim state-of-the-art visual quality across all
3D generation settings, but to isolate the efficiency--controllability--quality
trade-off obtained by moving diffusion into an explicit and compact geometric primitive space.

\textbf{Quality and diversity.}
Methods that diffuse dense 3D states still provide the strongest distribution matching. DiT-3D achieves near-ideal 1-NNA on \textsc{Chair} and \textsc{Car}, and the best COV on \textsc{Chair} and \textsc{Airplane}, reflecting the fidelity advantage of voxel-space denoising. In contrast, point-space diffusion methods (DPM, PVD, TIGER) denoise $2048\times3$ unconstrained coordinates and often suffer from larger distribution gaps, particularly on \textsc{Airplane}. LION narrows this gap by moving diffusion to a learned latent space, giving a more balanced 1-NNA/COV trade-off. DoSs occupies a different point in this spectrum: it sacrifices some dense-geometry fidelity for compactness and efficiency, while remaining competitive in distributional quality. It achieves strong 1-NNA, including the best result on \textsc{Airplane}, and high COV on \textsc{Chair}, though its COV on \textsc{Airplane} and \textsc{Car} still trails the strongest dense-state baseline.

% A primary driver of runtime is the dimensionality of the denoised state. Point diffusion operates on $\sim$24\,KB states ($2048\times3$ float32), voxel diffusion in DiT-3D uses a larger state ($32^3\times3$, $\sim$384\,KB), and mesh-hybrid diffusion can be larger still (e.g., $64^3\times4$, $\sim$4096\,KB). DoSs instead denoises $128\times15$ superquadric tokens, only $\sim$7\,KB, which translates into faster sampling in our setup (0.6\,s per shape at batch size 1; Tab.~\ref{tab:quant_main}). Overall, the results support our key claim: shifting diffusion from dense geometric fields to compact superquadric tokens yields a favorable efficiency--quality trade-off for unconditional generation at standard evaluation resolution. Some examples generated by DoSs are displayed in Fig.~\ref{fig:7}.
\textbf{Efficiency and diffusion state.}
Runtime is largely governed by the size of the denoised state. Point diffusion denoises $\sim$24\,KB states ($2048\times3$ float32), DiT-3D voxel diffusion uses $\sim$384\,KB states ($32^3\times3$), and mesh-hybrid diffusion can be even larger (e.g., $\sim$4096\,KB for $64^3\times4$). In contrast, DoSs denoises only $128\times15$ superquadric tokens, or $\sim$7\,KB. This compact state directly improves sampling efficiency, yielding 0.6\,s per shape at batch size 1 in our setup (Tab.~\ref{tab:quant_main}). These results support our central message: moving diffusion from dense geometric fields to compact superquadric tokens offers an effective efficiency--quality trade-off for unconditional 3D generation at standard evaluation resolution. Qualitative samples and denoising visualizations are shown in Fig.~\ref{fig:7} and Appendix~\ref{appendix:qv}.

% \paragraph{Multi-class training.}
\textbf{Multi-class training.} We further evaluate a single multi-class DoSs model trained jointly on
\textsc{Chair}, \textsc{Airplane}, and \textsc{Car} using category conditioning;
the detailed results are reported in Appendix~A. The joint model achieves
similar performance to the category-specific models in Tab.~\ref{tab:quant_main}
while requiring only 6 hours of training on one RTX 4090. This suggests that
DoSs is not restricted to category-specific training, although scaling to broader
and more diverse datasets remains future work.
% --------------------------------------------------------
\subsection{Experimental Analysis}
\label{sec:ablation}

This section reports ablation studies with \textsc{Chair} to investigate the benefits of key designs of our DoSs, including the existence token, the 6D orientation representation, and canonical sorting strategies. The result is shown in Tab.~\ref{tab:ablation_chair}.

\begin{table}[t]
\caption{Ablations on \textsc{Chair} under CD-based metrics. We ablate existence token, 6D orientation representation, and canonical sorting (volume-rank vs.\ $y$-rank).}
\label{tab:ablation_chair}
\centering
\small
\setlength{\tabcolsep}{4pt}
\renewcommand{\arraystretch}{1.05}
\begin{adjustbox}{max width=\columnwidth}
\begin{tabular}{cc cc cc}
\toprule
Exist. token & 6D orient. repr. & Vol.-rank & $y$-rank & 1-NNA $\downarrow$ & COV $\uparrow$ \\
\midrule
$\times$ & $\times$  & $\times$ & $\times$  & 93.32 & 19.91 \\
\checkmark & $\times$ & $\times$ & $\times$ & 58.36 & 48.41 \\
\checkmark & \checkmark & $\times$ & $\times$ & 57.57 & 50.37 \\
\checkmark & \checkmark & $\times$ & \checkmark & 55.28 & 50.93 \\
\checkmark & \checkmark & \checkmark & $\times$ & \textbf{53.80} & \textbf{51.31} \\
\bottomrule
\end{tabular}
\end{adjustbox} \vspace{-5mm}
\end{table} 

\textbf{Existence token.}
We ablate our design of existence token by comparing (i) with existance score ($e_k$ in Eq.~\ref{eq:ddpm-reverse}) vs. (ii) without existance score. In this case, the model directly predicts the set of 128 superquadrics for each object. During training, we first shuffled the order of superquadrics in each training data. Then we padded the rest rows by copying existing superquadrics. The result indicates that the existence token is the design that contributes the most to the performance of our model. 
Comparing the first two rows of Tab.~\ref{tab:ablation_chair}, the existence token significantly improves performance (by $34.96\%$ @ 1-NNA and $28.50\%$ @ COV). 
This indicates that explicitly modeling the presence or absence of valid superquadric primitives is crucial.
% We ablate our pose modeling strategy for each primitive, comparing (i) a naive unconstrained parameterization vs.\ (ii) an SE(3)-aware factorization (translation + rotation) used in our final model.

\textbf{Rotation representation.}
We compare (i) Euler angles vs. (ii) continuous 6D rotation representation~\cite{zhou2019continuity}, which avoids discontinuities and improves optimization stability. The result shows that adding the 6D rotation representation slightly improves the model performance (by $0.79\%$ @ 1-NNA and $1.96\%$ @ COV), which confirms that the continuous 6D representation stabilizes rotation modeling by avoiding discontinuities inherent in Euler angles. However, the performance of the 6D representation and Euler angles is comparable. We believe this may be due to the fact that, when Euler angles are constrained within a single period (e.g., from $- \pi$ to $\pi$), discontinuities in Euclidean space occur only at the boundaries, which are rarely encountered in our case.

\textbf{Canonical sorting.}
In this section, we study three ordering strategies:
(i) random order, (ii) sorting primitives by decreasing volume, and (iii) sorting by vertical position ($y$-axis) to align with gravity-consistent structure in chairs. As shown in the last three rows, position based sorting reduces 1-NNA from $57.57\%$ to $55.28\%$ and increases COV from $50.37\%$ to $50.93\%$. Meanwhile, volume-based sorting reduces 1-NN accuracy from $57.57\%$ to $53.80\%$ and increases coverage from $50.37\%$ to $51.13\%$. This suggests a consistent and geometrically meaningful ordering is beneficial to the model performance and sorting superquadric by volume can yield better results for the final outcome.

\subsection{Capabilities Enabled by DoSs}
\label{sec:tasks}

We showcase three capabilities that highlight the scalability and controllability of DoSs in a compact, explicit space.

\begin{figure}[t]
\centering
\setlength{\tabcolsep}{2pt}
\begin{tabular}{ccc}
\includegraphics[width=0.31\columnwidth]{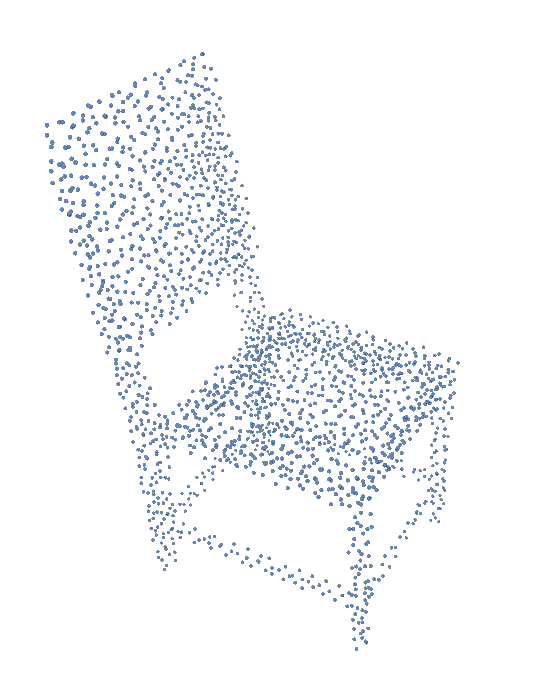} &
\includegraphics[width=0.31\columnwidth]{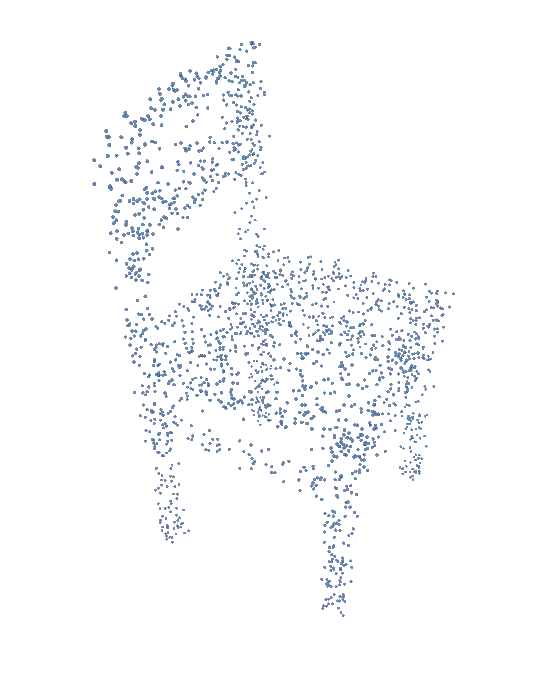} &
\includegraphics[width=0.31\columnwidth]{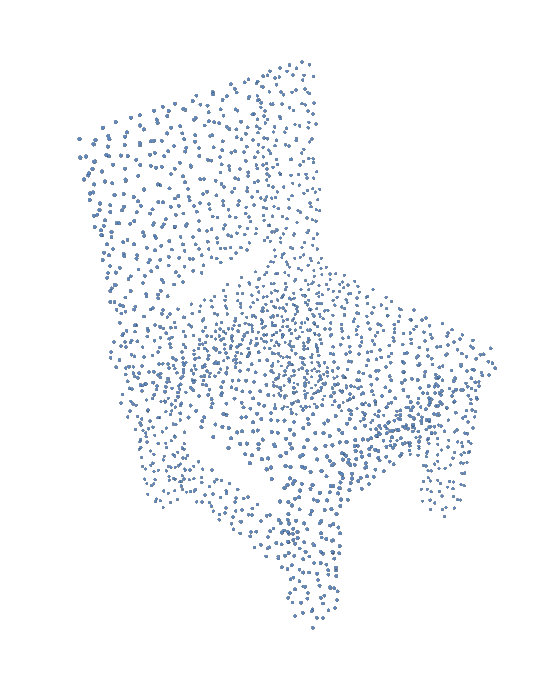} \\
{\small (a) GT} & {\small (b) DiT-3D (1-NN)} & {\small (c) Ours (1-NN)} \\
\end{tabular}
\vspace{-2mm}
\captionof{figure}{Resolution free point decoding on \textsc{Chair}. We decode generated outputs into 2{,}048 surface points for visualization and evaluation. DoSs produces a more complete part structure than DiT-3D in this example, while CD-based distributional metrics still favor voxel diffusion spaces due to their closer match to point-level detail and sampling statistics.}
\label{fig:nn_sampling_compare}
\vspace{-6mm}
\end{figure}

\subsubsection{Resolution Free Point Decoding}
\label{sec:tasks:resfree}

DoSs predicts superquadric tokens, which we decode into point clouds using the sampling pipeline in Sec.~\ref{sec:method:resfree}: we sample surface points for each predicted primitive, remove points that lie inside other primitives to handle overlaps, and apply FPS to obtain a point cloud of any requested size $N$. Fig.~\ref{fig:nn_sampling_compare} shows an example on \textsc{Chair} (2{,}048 points). DiT-3D exhibits missing or deformed regions, while DoSs yields a more complete and coherent part layout. This is expected because DoSs operates in an explicit primitive space: each token represents a closed, smooth geometric part with structured degrees of freedom (shape, size, pose), which acts as a strong prior during generation.

Better visual plausibility does not necessarily imply better 1-NNA/COV under CD. These metrics depend only on nearest neighbor distances between point clouds, and CD is sensitive to point placement, density, and fine-scale details. Our decoding enforces a specific surface sampling distribution and overlap filtering, which can increase CD even when global structure looks cleaner. In addition, primitive tokens can bias samples toward smoother, more regular shapes, which may reduce fine-grained diversity and lower COV.

\subsubsection{Structure Aware Editability}
\label{sec:tasks:edit}

DoSs generates shape as a compact set of explicit superquadric tokens, so edits can be applied directly in token space using the operations in Sec.~\ref{sec:method:edit} (part deletion via the existence channel, part scaling via axis lengths, part reposing via pose parameters, and part reshaping via the shape exponents). Fig.~\ref{fig:teaser} shows \textsc{Chair} edits produced by these simple parameter level manipulations. In contrast, dense diffusion spaces (points, voxels, meshes) and latent or implicit representations do not expose stable part level parameters. Achieving comparable edits typically requires additional conditioning, guidance, or post processing. Their fine grained, localized control is harder because the edit target is entangled across many coordinates or latent variables rather than a small number of interpretable part parameters.

\begin{figure}[t]
\centering
\setlength{\tabcolsep}{1pt}
\begin{tabular}{ccc}
\includegraphics[width=0.27\columnwidth]{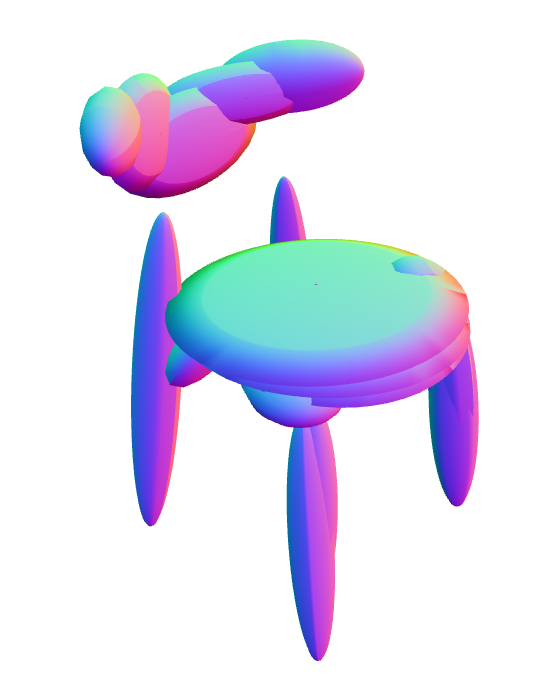} &
\includegraphics[width=0.27\columnwidth]{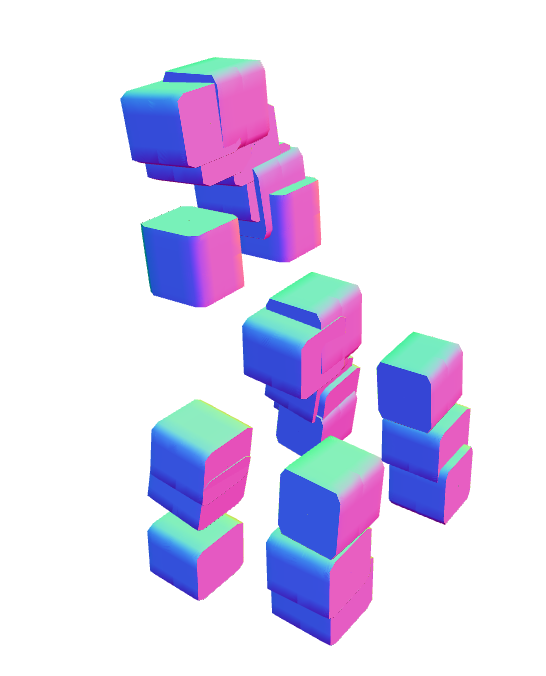} &
\includegraphics[width=0.27\columnwidth]{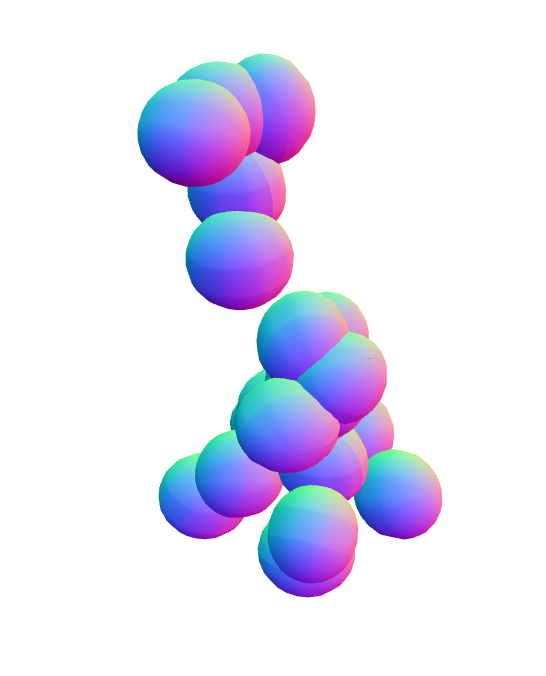} \\
{\scriptsize (a) $\epsilon_1{=}\epsilon_2{=}1$} &
{\scriptsize (b) $\epsilon_1{=}\epsilon_2{=}0,\mathbf{s}_k=const$} &
{\scriptsize (c) $\epsilon_1{=}\epsilon_2{=}1,\mathbf{s}_k=const$}
\end{tabular}
\vspace{-2mm}
\caption{
Failure modes of constrained denoising under overly restrictive constraints.
Fixing exponents (a) already narrows the feasible shape family; additionally clamping all three size axes (b,c), i.e., $\mathbf{s}_k=const$, can over-constrain the process, leading to degenerate or implausible primitive assemblies.
}
\label{fig:design_fail}
\vspace{-6mm}
\end{figure}

\subsubsection{Geometric Aware Design}
\label{sec:tasks:design}

DoSs supports geometric aware design by constraining interpretable token dimensions during sampling, and only denoising the remaining free variables. Following constrained denoising ideas such as DPS~\citep{chung2022diffusion}, our design interface in Sec.~\ref{sec:method:design} enforces simple hard constraints via per step clamping or projection in token space. For example, to design a chair from bamboo like parts with fixed radius, we constrain each primitive to be cylinder like by setting the shape exponents to a target pair (e.g., $\epsilon_{k1}\!\approx\!0$, $\epsilon_{k2}\!=\!1$) and fixing the cross section $a_{k1}\!=\!a_{k2}\!=r$, while denoising the remaining variables (height $a_{k3}$, pose, and existence). Similarly, a plank only design is obtained by fixing a cuboid like shape and clamping one axis length as the thickness. Fig.~\ref{fig:teaser} shows representative constrained generations.

This capability is difficult to realize in dense or implicit diffusion spaces, where constraints must be expressed over thousands to millions of coupled variables (voxels, points, or latents), typically requiring additional guidance or post processing, and fine grained part level control is much less direct. However, constrained denoising also has limitations. If the constraints are overly restrictive, the feasible shape family can become too small for the learned prior, causing degenerate or implausible primitive assemblies. Fig.~\ref{fig:design_fail} shows typical failure modes when clamping exponents and, more severely, clamping all size axes.

\section{Limitations and Future Works}

\textbf{Limitations.}
Diffusion over superquadrics (DoSs) trades representational capacity for compactness and interpretability.
As a result, with a bounded number of primitives, the model is biased toward piecewise-smooth, low-frequency geometry and can under-represent thin structures, sharp edges, deep concavities, and topology-heavy details that are easier to express in dense fields or meshes.
In addition, DoSs inherits the failure modes of the superquadrics fitting methods: imperfect fitting introduces structured label noise (e.g., missed parts and unstable poses), while non-identifiability of parameters (multiple equivalent superquadrics parameterizations for nearly identical surfaces) makes the target distribution in token space more multimodal than in surface space, which can hinder stable learning. Finally, our evaluation relies on CD-based 1-NNA/COV computed on point clouds sampled from decoded surfaces. These Chamfer-based metrics can be misaligned with visual quality and may under-reflect our gains; more broadly, unified 3D generation metrics remain an open problem.

\textbf{Future Work.}
The limitations and strengths of DoSs suggest two promising directions: improving fine geometric detail and advancing physically grounded generation. A central challenge in 3D diffusion is to reconcile object-level structure, such as part layout, symmetry, and functional organization, with surface-level detail, such as thin elements, sharp boundaries, and high-frequency geometry. DoSs deliberately emphasizes coherent global structure by diffusing a compact set of primitive parameters, but this abstraction can limit local geometric fidelity. To improve visual quality, we will explore more accurate superquadric fitting/decomposition, more expressive primitive variants, and improved parameter identifiability. We will also study hybrid pipelines where DoSs generates a structurally consistent coarse shape that is later refined by higher-capacity representations, such as local implicit patches, residual point refinement, or mesh/field upsampling.

In parallel, we will extend DoSs toward dynamic and physically grounded settings, where efficiency, feasibility, and editability are often as important as surface realism. Examples include simulation-valid generation, collision/contact reasoning, stability constraints, articulated or part-aware objects, and large-scale world modeling with long-horizon generation. DoSs is well suited to these applications: its explicit primitives make constraints such as non-penetration, support, symmetry, contact, and stability easy to impose and verify, while its compact token space enables fast sampling and iterative feasibility correction without the heavy cost of dense 3D diffusion.

\section{Conclusion}
We presented DoSs, which shifts diffusion from dense geometric representations to a compact and explicit space of superquadric tokens.
By denoising explicit primitive parameters (pose, scale, and shape) rather than thousands-to-millions of dense values, DoSs substantially reduces the diffusion-state dimensionality while retaining the ability to decode 3D surfaces and point clouds.
Across standard benchmarks, DoSs achieves competitive surface fidelity and distributional quality after decoding to point clouds, while offering shorter training and inference times.
Moreover, DoSs enables structure-aware capabilities that are difficult to realize in dense diffusion spaces, including resolution-free sampling, part-level editability, and constrained denoising for geometry-aware design.
Overall, DoSs suggests that diffusing in an explicit and compact primitive space is a practical route toward scalable and controllable 3D generation. Future work includes improving tokenization and denoisers for higher fidelity, most promisingly, leveraging DoSs' compact and explicit tokens for efficient multimodal/scene conditioning and for physically grounded, temporally structured generation with geometric constraints.

\section*{Impact Statement}
This paper aims to advance 3D generative modeling by moving the diffusion process from dense geometric representations to compact and explicit superquadric tokens. By reducing the dimensionality of the generative space, our approach can lower compute and memory costs of training and sampling, which may reduce energy use and improve accessibility of 3D generation research and applications for groups with limited hardware. The explicit, part-based parameterization can also support more interpretable and controllable 3D synthesis, which may benefit downstream uses such as simulation, robotics, AR/VR content creation, and rapid prototyping.

As with many general-purpose 3D generation techniques, there are potential risks. Generated shapes could be repurposed to aid the design of harmful or unsafe objects, or be used in contexts where fabricated parts require strict safety certification. Moreover, training data may encode biases in object categories or styles, leading to uneven performance or stereotyped outputs; and generated shapes may inadvertently resemble copyrighted or proprietary designs if such patterns exist in the data. To mitigate these concerns, we encourage responsible release and use practices (e.g., adherence to dataset and model licenses, clear documentation of intended use and limitations, and optional filtering of sensitive categories where applicable). We also emphasize that outputs from this model are not validated for safety-critical deployment and should undergo application-specific review, testing, and compliance checks before real-world use.

\section*{Acknowledgement}
We thank anonymous ICML reviewers for their useful feedback and suggestions. 

This research was supported in part by the Advanced Robotics Centre (ARC) of the National University of Singapore.

\nocite{langley00}

\bibliography{reference}

@inproceedings{langley00,
 author    = {P. Langley},
 title     = {Crafting Papers on Machine Learning},
 year      = {2000},
 pages     = {1207--1216},
 editor    = {Pat Langley},
 booktitle     = {Proceedings of the 17th International Conference
              on Machine Learning (ICML 2000)},
 address   = {Stanford, CA},
 publisher = {Morgan Kaufmann}
}

@inproceedings{zhou2019continuity,
  title     = {On the Continuity of Rotation Representations in Neural Networks},
  author    = {Zhou, Yi and Barnes, Connelly and Lu, Jingwan and Yang, Jimei and Li, Hao},
  booktitle = {Proceedings of the IEEE/CVF Conference on Computer Vision and Pattern Recognition (CVPR)},
  year      = {2019},
  pages     = {5738--5746},
  doi       = {10.1109/CVPR.2019.00589},
  eprint    = {1812.07035},
  archivePrefix = {arXiv},
  primaryClass  = {cs.LG},
  url       = {https://arxiv.org/abs/1812.07035}
}

@inproceedings{ren2024tiger,
  title     = {TIGER: Time-Varying Denoising Model for 3D Point Cloud Generation with Diffusion Process},
  author    = {Ren, Zhiyuan and Kim, Minchul and Liu, Feng and Liu, Xiaoming},
  booktitle = {Proceedings of the IEEE/CVF Conference on Computer Vision and Pattern Recognition (CVPR)},
  year      = {2024}
}

@inproceedings{peebles2023scalable,
  title={Scalable diffusion models with transformers},
  author={Peebles, William and Xie, Saining},
  booktitle={Proceedings of the IEEE/CVF international conference on computer vision},
  pages={4195--4205},
  year={2023}
}

@inproceedings{bao2023one,
  title={One transformer fits all distributions in multi-modal diffusion at scale},
  author={Bao, Fan and Nie, Shen and Xue, Kaiwen and Li, Chongxuan and Pu, Shi and Wang, Yaole and Yue, Gang and Cao, Yue and Su, Hang and Zhu, Jun},
  booktitle={International Conference on Machine Learning},
  pages={1692--1717},
  year={2023},
  organization={PMLR}
}

@inproceedings{zhou20213d,
  title={3d shape generation and completion through point-voxel diffusion},
  author={Zhou, Linqi and Du, Yilun and Wu, Jiajun},
  booktitle={Proceedings of the IEEE/CVF international conference on computer vision},
  pages={5826--5835},
  year={2021}
}

@inproceedings{zeng2022lion,
  title={LION: Latent Point Diffusion Models for 3D Shape Generation},
  author={Zeng, Xiaohui and Vahdat, Arash and Williams, Francis and Gojcic, Zan and Litany, Or and Fidler, Sanja and Kreis, Karsten},
  booktitle={Advances in Neural Information Processing Systems},
  year={2022}
}

@article{liu2023meshdiffusion,
  title={Meshdiffusion: Score-based generative 3d mesh modeling},
  author={Liu, Zhen and Feng, Yao and Black, Michael J and Nowrouzezahrai, Derek and Paull, Liam and Liu, Weiyang},
  journal={arXiv preprint arXiv:2303.08133},
  year={2023}
}

@article{song2020score,
  title={Score-based generative modeling through stochastic differential equations},
  author={Song, Yang and Sohl-Dickstein, Jascha and Kingma, Diederik P and Kumar, Abhishek and Ermon, Stefano and Poole, Ben},
  journal={arXiv preprint arXiv:2011.13456},
  year={2020}
}

@article{chang2015shapenet,
  title={Shapenet: An information-rich 3d model repository},
  author={Chang, Angel X and Funkhouser, Thomas and Guibas, Leonidas and Hanrahan, Pat and Huang, Qixing and Li, Zimo and Savarese, Silvio and Savva, Manolis and Song, Shuran and Su, Hao and others},
  journal={arXiv preprint arXiv:1512.03012},
  year={2015}
}

@inproceedings{luo2021diffusion,
  title={Diffusion probabilistic models for 3d point cloud generation},
  author={Luo, Shitong and Hu, Wei},
  booktitle={Proceedings of the IEEE/CVF conference on computer vision and pattern recognition},
  pages={2837--2845},
  year={2021}
}

@article{mo2023dit,
  title={Dit-3d: Exploring plain diffusion transformers for 3d shape generation},
  author={Mo, Shentong and Xie, Enze and Chu, Ruihang and Hong, Lanqing and Niessner, Matthias and Li, Zhenguo},
  journal={Advances in neural information processing systems},
  volume={36},
  pages={67960--67971},
  year={2023}
}

@inproceedings{kalischek2024tetradiffusion,
  title={TetraDiffusion: Tetrahedral diffusion models for 3D shape generation},
  author={Kalischek, Nikolai and Peters, Torben and Wegner, Jan D and Schindler, Konrad},
  booktitle={European Conference on Computer Vision},
  pages={357--373},
  year={2024},
  organization={Springer}
}

@inproceedings{xiong2025octfusion,
  title={OctFusion: Octree-based Diffusion Models for 3D Shape Generation},
  author={Xiong, Bojun and Wei, Si-Tong and Zheng, Xin-Yang and Cao, Yan-Pei and Lian, Zhouhui and Wang, Peng-Shuai},
  booktitle={Computer Graphics Forum},
  volume={44},
  number={5},
  pages={e70198},
  year={2025},
  organization={Wiley Online Library}
}

@inproceedings{ren2024xcube,
  title={Xcube: Large-scale 3d generative modeling using sparse voxel hierarchies},
  author={Ren, Xuanchi and Huang, Jiahui and Zeng, Xiaohui and Museth, Ken and Fidler, Sanja and Williams, Francis},
  booktitle={Proceedings of the IEEE/CVF conference on computer vision and pattern recognition},
  pages={4209--4219},
  year={2024}
}

@article{wu2024direct3d,
  title={Direct3d: Scalable image-to-3d generation via 3d latent diffusion transformer},
  author={Wu, Shuang and Lin, Youtian and Zhang, Feihu and Zeng, Yifei and Xu, Jingxi and Torr, Philip and Cao, Xun and Yao, Yao},
  journal={Advances in Neural Information Processing Systems},
  volume={37},
  pages={121859--121881},
  year={2024}
}

@article{wang2025diffusion,
  title={Diffusion models for 3D generation: A survey},
  author={Wang, Chen and Peng, Hao-Yang and Liu, Ying-Tian and Gu, Jiatao and Hu, Shi-Min},
  journal={Computational Visual Media},
  volume={11},
  number={1},
  pages={1--28},
  year={2025},
  publisher={TUP}
}

@article{fedele2025superdec,
  title={Superdec: 3d scene decomposition with superquadric primitives},
  author={Fedele, Elisabetta and Sun, Boyang and Guibas, Leonidas and Pollefeys, Marc and Engelmann, Francis},
  journal={arXiv preprint arXiv:2504.00992},
  year={2025}
}

@article{zuo2025quadricformer,
  title={QuadricFormer: Scene as Superquadrics for 3D Semantic Occupancy Prediction},
  author={Zuo, Sicheng and Zheng, Wenzhao and Han, Xiaoyong and Yang, Longchao and Pan, Yong and Lu, Jiwen},
  journal={arXiv preprint arXiv:2506.10977},
  year={2025}
}

@article{nichol2022point,
  title={Point-e: A system for generating 3d point clouds from complex prompts},
  author={Nichol, Alex and Jun, Heewoo and Dhariwal, Prafulla and Mishkin, Pamela and Chen, Mark},
  journal={arXiv preprint arXiv:2212.08751},
  year={2022}
}

@article{wu2023sin3dm,
  title={Sin3dm: Learning a diffusion model from a single 3d textured shape},
  author={Wu, Rundi and Liu, Ruoshi and Vondrick, Carl and Zheng, Changxi},
  journal={arXiv preprint arXiv:2305.15399},
  year={2023}
}

@article{barr1981superquadrics,
  title={Superquadrics and angle-preserving transformations},
  author={Barr, Alan H},
  journal={IEEE Computer graphics and Applications},
  volume={1},
  number={01},
  pages={11--23},
  year={1981},
  publisher={IEEE Computer Society}
}

@inproceedings{alaniz2023iterative,
  title={Iterative superquadric recomposition of 3d objects from multiple views},
  author={Alaniz, Stephan and Mancini, Massimiliano and Akata, Zeynep},
  booktitle={Proceedings of the IEEE/CVF International Conference on Computer Vision},
  pages={18013--18023},
  year={2023}
}

@article{monnier2023differentiable,
  title={Differentiable blocks world: Qualitative 3d decomposition by rendering primitives},
  author={Monnier, Tom and Austin, Jake and Kanazawa, Angjoo and Efros, Alexei and Aubry, Mathieu},
  journal={Advances in Neural Information Processing Systems},
  volume={36},
  pages={5791--5807},
  year={2023}
}

@book{jaklic2000segmentation,
  title={Segmentation and recovery of superquadrics},
  author={Jaklic, Ales and Leonardis, Ales and Solina, Franc},
  volume={20},
  year={2000},
  publisher={Springer Science \& Business Media}
}

@inproceedings{sohl2015deep,
  title={Deep unsupervised learning using nonequilibrium thermodynamics},
  author={Sohl-Dickstein, Jascha and Weiss, Eric and Maheswaranathan, Niru and Ganguli, Surya},
  booktitle={International conference on machine learning},
  pages={2256--2265},
  year={2015},
  organization={pmlr}
}

@article{ho2020ddpm,
  title={Denoising diffusion probabilistic models},
  author={Ho, Jonathan and Jain, Ajay and Abbeel, Pieter},
  journal={Advances in neural information processing systems},
  volume={33},
  pages={6840--6851},
  year={2020}
}

@inproceedings{liu2022robust,
  title={Robust and accurate superquadric recovery: A probabilistic approach},
  author={Liu, Weixiao and Wu, Yuwei and Ruan, Sipu and Chirikjian, Gregory S},
  booktitle={Proceedings of the IEEE/CVF Conference on Computer Vision and Pattern Recognition},
  pages={2676--2685},
  year={2022}
}

@article{chung2022diffusion,
  title={Diffusion posterior sampling for general noisy inverse problems},
  author={Chung, Hyungjin and Kim, Jeongsol and Mccann, Michael T and Klasky, Marc L and Ye, Jong Chul},
  journal={arXiv preprint arXiv:2209.14687},
  year={2022}
}

@inproceedings{liu2023marching,
  title={Marching-primitives: Shape abstraction from signed distance function},
  author={Liu, Weixiao and Wu, Yuwei and Ruan, Sipu and Chirikjian, Gregory S},
  booktitle={Proceedings of the IEEE/CVF Conference on Computer Vision and Pattern Recognition},
  pages={8771--8780},
  year={2023}
}

@inproceedings{wu2023fast,
  title={Fast point cloud generation with straight flows},
  author={Wu, Lemeng and Wang, Dilin and Gong, Chengyue and Liu, Xingchao and Xiong, Yunyang and Ranjan, Rakesh and Krishnamoorthi, Raghuraman and Chandra, Vikas and Liu, Qiang},
  booktitle={Proceedings of the IEEE/CVF conference on computer vision and pattern recognition},
  pages={9445--9454},
  year={2023}
}
\bibliographystyle{icml2026}

%%%%%%%%%%%%%%%%%%%%%%%%%%%%%%%%%%%%%%%%%%%%%%%%%%%%%%%%%%%%%%%%%%%%%%%%%%%%%%%
%%%%%%%%%%%%%%%%%%%%%%%%%%%%%%%%%%%%%%%%%%%%%%%%%%%%%%%%%%%%%%%%%%%%%%%%%%%%%%%
% APPENDIX
%%%%%%%%%%%%%%%%%%%%%%%%%%%%%%%%%%%%%%%%%%%%%%%%%%%%%%%%%%%%%%%%%%%%%%%%%%%%%%%
%%%%%%%%%%%%%%%%%%%%%%%%%%%%%%%%%%%%%%%%%%%%%%%%%%%%%%%%%%%%%%%%%%%%%%%%%%%%%%%
\newpage
\appendix
\onecolumn
\section{Results on Multi-class Training}
\label{multiclass}
\begin{table}[H]
\centering
\caption{Quantitative evaluation on different object categories.}
\label{tab:category_results}
\small
\begin{tabular}{lcc}
\toprule
Category & 1-NNA-CD  & COV-CD  \\
\midrule
Chair     & 55.70 & 50.19  \\
Airplane  & 60.20 & 45.36  \\
Car       & 61.11 & 47.78  \\
\bottomrule
\end{tabular}
\vspace{-5mm}
\end{table}

We also trained DoSs in {\textsc{Chair}, \textsc{Car}, \textsc{Airplane}} together to show the effectiveness of our model on multi-class training. 
Since our model is unconditional and generating objects of a specific type requires taking the category information as condition, we modified our model by encoding the category index with 3-layer MLP, concatenating with time embedding and input to the noise prediction network in DDPM. 
We evaluated the multi-class model with 1000 objects of each category, the result of 1-NNA and COV is illustrated in Tab.~\ref{tab:category_results}. 
We can observe that the DoSs achieves competitive generation results against category-specific models for all metrics.
This result demonstrates that DoSs hold the potential to simultaneously generate objects of all categories by training a single global model.
The multi-categories model is trained with around 8{,}000 objects for 6 hours on one RTX 4090. 

\section{Primitive Budget Analysis}
\label{app:primitive_budget}

DoSs represents each shape using a fixed token budget of \(K_{\max}=128\)
superquadric slots, while the actual number of active primitives varies across
objects and is controlled by the existence score. To justify this choice, we
report the statistics of the superquadric decompositions used in our ShapeNet
experiments in Tab.~\ref{tab:sq_count_stats}.

\begin{table}[h]
\centering
\caption{
Statistics of the number of superquadrics per shape in the extracted
ShapeNet superquadrics. P90 denotes the 90th percentile: 90\% of shapes in
that category use no more than this number of primitives.
}
\label{tab:sq_count_stats}
\begin{tabular}{lrrrr}
\toprule
Category & Mean & Median & P90 & Max \\
\midrule
\textsc{Airplane} & 12.98 & 13 & 17 & 52 \\
\textsc{Chair}    & 21.08 & 19 & 36 & 123 \\
\textsc{Car}      & 18.64 & 13 & 38 & 100 \\
\bottomrule
\end{tabular}
\end{table}

The statistics show that \(K_{\max}=128\) is sufficient for the current
ShapeNet setting. All observed decompositions fit within this budget, with
maximum primitive counts of 52, 123, and 100 for \textsc{Airplane},
\textsc{Chair}, and \textsc{Car}, respectively. Moreover, the P90 values are
much smaller than 128, indicating that most shapes use only a small fraction of
the available slots. This suggests that, for the evaluated categories, the
fixed primitive budget is not the main bottleneck.

Increasing the budget to larger states such as \(256 \times 15\) or
\(512 \times 15\) may improve representational capacity for more complex
objects, but it would also increase the diffusion-state size and sampling cost.
For the current ShapeNet benchmarks, we therefore use \(K_{\max}=128\) as a
compact budget that covers the extracted primitives while keeping
the denoising state small. Exploring larger or flexible-length primitive
generation, such as token-by-token autoregressive superquadric generation,
is an interesting direction for future work on more diverse datasets such as
Objaverse.

\section{Parameter Sensitivity Analysis}
\label{app:param_sensitivity}

DoSs denoises heterogeneous superquadric parameters, including position,
rotation, size, shape exponents, and existence scores. These parameters have
different physical meanings and may affect the denoising process differently.
To better understand this behavior, we conduct a perturbation study
by perturbing one parameter group at a time during sampling.

For a fixed sampling seed, we first generate a reference output
\(\tilde{Z}_0\). We then repeat sampling
with the same seed, but add random perturbation to only one parameter group
while keeping the remaining groups unchanged. Specifically, for each group
\(g \in \{\text{shape}, \text{size}, \text{rotation}, \text{position},
\text{existence}\}\), we add \(10\%\) random noise to the corresponding
normalized token dimensions during denoising. After sampling, we compare the
perturbed output \(\tilde{Z}^{(g)}_0\) with the reference output using the
Manhattan distance: $\Delta_g = \left\| \tilde{Z}^{(g)}_0 - \tilde{Z}_0 \right\|_1 .$
Since different parameter groups have different dimensionalities, we also
report the deviation per dimension, computed as \(\Delta_g / d_g\), where
\(d_g\) is the dimensionality of the perturbed group.

\begin{table}[h]
\centering
\caption{
Parameter sensitivity under group-wise perturbations during denoising.
For each parameter group, we add \(10\%\) random noise to that group only and
measure the final deviation from the unperturbed output using Manhattan
distance. The per-dimension deviation accounts for different group
dimensionalities.
}
\label{tab:param_sensitivity}
\begin{tabular}{lrrr}
\toprule
Parameter group & Dim. & Total deviation & Deviation / dim. \\
\midrule
Shape     & 2 & 153.90 & 76.95  \\
Size      & 3 & 104.20 & 34.73  \\
Rotation  & 6 & 138.78 & 23.13  \\
Position  & 3 & 132.40 & 44.13  \\
Existence & 1 & 101.59 & 101.59 \\
\bottomrule
\end{tabular}
\end{table}

The results suggest that the denoising process is most sensitive to coarse
structural factors. In particular, the existence score has the largest
per-dimension deviation, which is consistent with its role in determining
whether a primitive is active. Shape parameters are also highly influential,
as they control the primitive family and therefore strongly affect the global
geometry. Position perturbations produce a non-negligible deviation because
they directly change the spatial layout of parts. In contrast, size and
rotation have smaller per-dimension deviations in this diagnostic study.

This analysis should be interpreted as a preliminary robustness diagnostic
rather than a complete perceptual evaluation. The reported distances are
measured in the normalized token space, not directly in surface space.
Nevertheless, the trend supports the importance of explicitly modeling
existence and shape parameters, and helps explain why the existence-token
design contributes strongly to the performance observed in
Tab.~\ref{tab:ablation_chair}.

\section{Discussion}
\label{app:discussion}

\textbf{Scaling diffusion in 3D: where does the compute really go?}
A central challenge in 3D diffusion is that iterative denoising amplifies the cost of the chosen representation: even modest increases in spatial resolution or point count can multiply per-step compute/memory, and the cumulative cost over dozens to hundreds of steps becomes the bottleneck.
DoSs directly targets this challenge by moving diffusion to a compact, continuous token space, making long denoising chains and repeated sampling (e.g., for selection, reranking, or constrained generation) substantially more practical.
This reframes a common question in 3D diffusion: \emph{how to afford iterative refinement in high-dimensional spaces}, into a representation choice: reducing state dimensionality can unlock additional algorithmic degrees of freedom (e.g., stronger conditioning, constraint projection, or multi-try sampling) that are otherwise prohibitively expensive.

\textbf{Controllability and constraints: from ``prompting'' to verifiable geometry.}
A major limitation of many 3D diffusion systems is that control is often indirect: users condition on text or images but lack an explicit interface to enforce geometric properties.
DoSs offers a different control primitive: constraints can be expressed directly on interpretable parameters (pose/scale/shape) and checked against decoded geometry (e.g., collision, support, symmetry).
This suggests a broader shift in how we think about controllable 3D diffusion: rather than relying solely on learned conditioning signals, we can incorporate \emph{explicit constraint operators} during sampling.
The open challenge is to make such constraints robust (avoid over-constraining and mode collapse) while remaining expressive (support multiple feasible solutions); compact token spaces make iterative constraint enforcement and multi-hypothesis sampling substantially more feasible.

\textbf{Evaluation: decoding bias, metric sensitivity, and the open problem of unified benchmarks.}
While point-cloud metrics such as CD/EMD-based 1-NNA and COV are widely used, they entangle the generative model with the \emph{decoder and sampling procedure}.
For token-based methods, small changes in surface sampling (density, stratification, noise, coverage of thin structures) can noticeably affect distance estimates, and thus change 1-NNA/COV even when rendered geometry looks similar.
Moreover, distributional metrics can disagree with perceptual quality: a method may generate visually plausible shapes that are nonetheless ``easy to separate'' under nearest-neighbor tests if it induces systematic biases (e.g., smoother surfaces, reduced micro-variations, or a narrower manifold after decoding).
These issues highlight an open problem for the community: establishing \emph{unified, representation-agnostic evaluation} that remains stable across decoding pipelines and captures both structural plausibility and geometric fidelity.
% In the near term, a more reliable practice is to report a \emph{suite} of metrics and protocols: (i) multiple distributional scores under standardized sampling and normalization, (ii) complementary measures that probe structure (e.g., part count statistics or symmetry consistency when available), and (iii) downstream/task-based evaluations when the application is known (e.g., simulation feasibility, collision rates, or grasp/pose success proxies).
A unified metric that is simultaneously faithful, stable across representations, and aligned with human judgment remains an important open question in 3D generation.

\section{Qualitative Visualization}
\label{appendix:qv}

\begin{figure*}
\centering
  \includegraphics[width=0.9\textwidth]{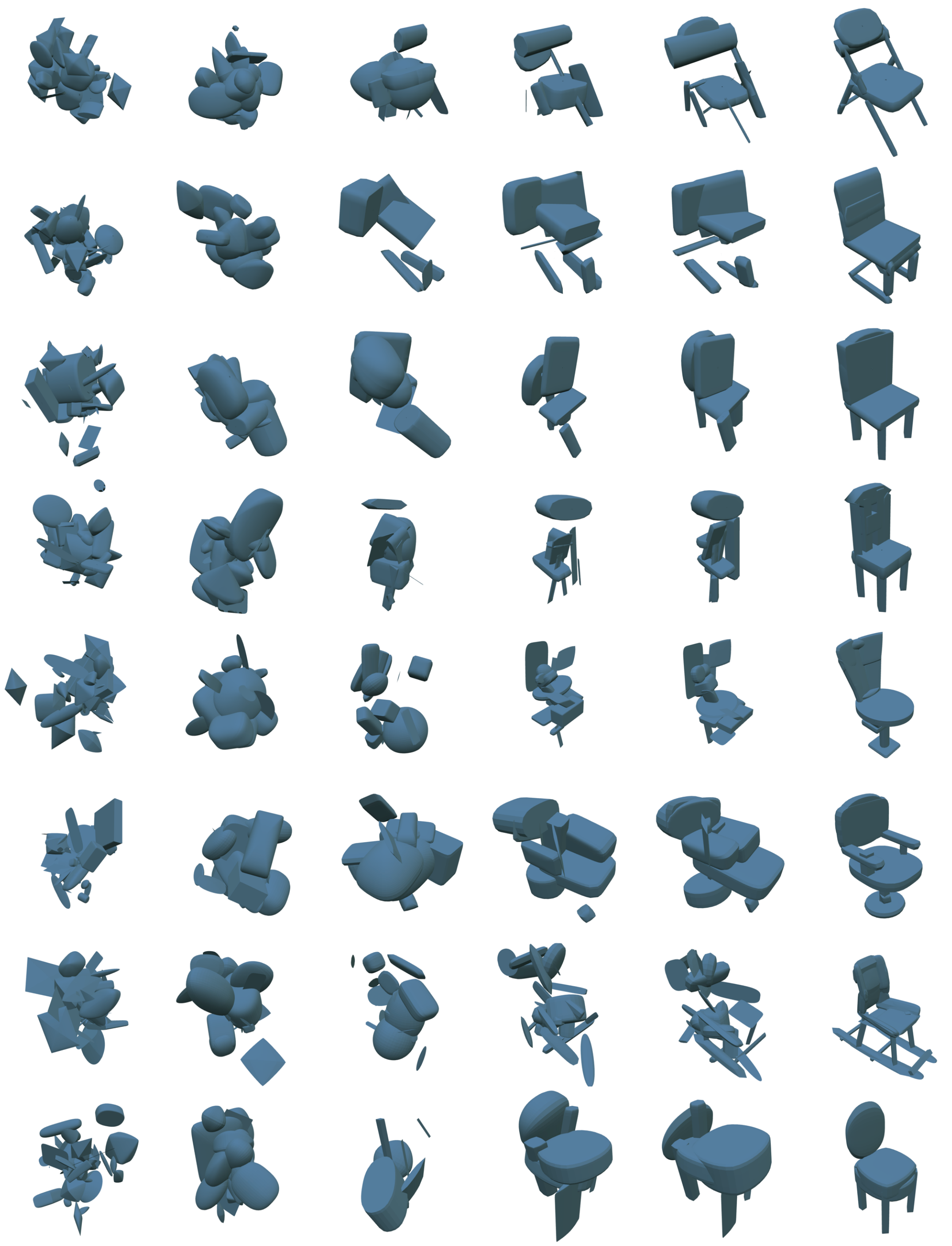}
  \captionof{figure}{Qualitative visualizations of the denoising process on \textsc{Chair} shape generation. The results of generating from random noise to final 3D shapes are shown in left-to-right order.}
  \label{fig:1}
\end{figure*}

\begin{figure*}
\centering
  \includegraphics[width=0.9\textwidth]{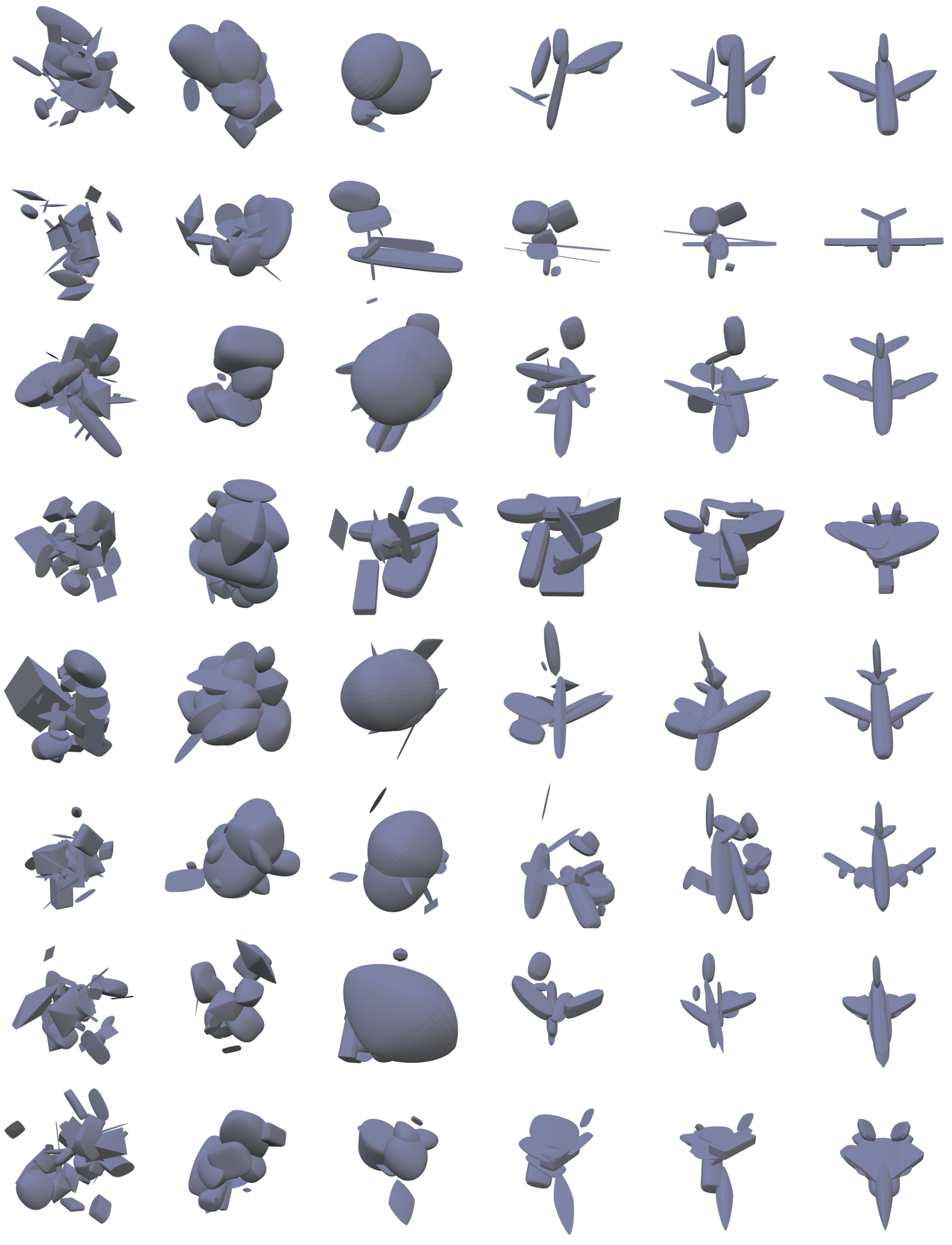}
  \captionof{figure}{Qualitative visualizations of the denoising process on \textsc{Airplane} shape generation. The results of generating from random noise to final 3D shapes are shown in left-to-right order.}
  \label{fig:2}
\end{figure*}

\begin{figure*}
\centering
  \includegraphics[width=0.9\textwidth]{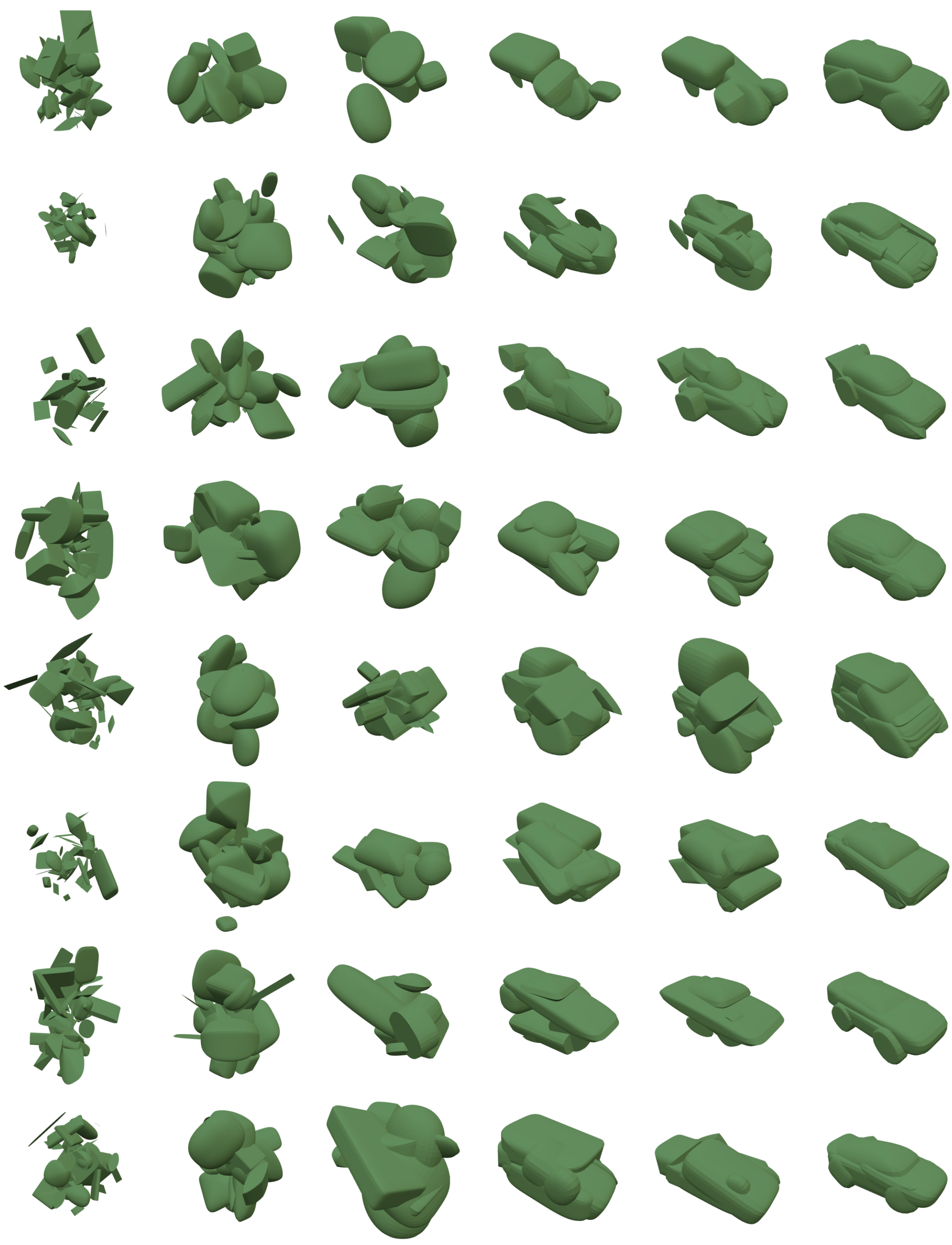}
  \captionof{figure}{Qualitative visualizations of the denoising process on \textsc{Car} shape generation. The results of generating from random noise to final 3D shapes are shown in left-to-right order.}
  \label{fig:3}
\end{figure*}

\begin{figure*}
\centering
  \includegraphics[width=0.9\textwidth]{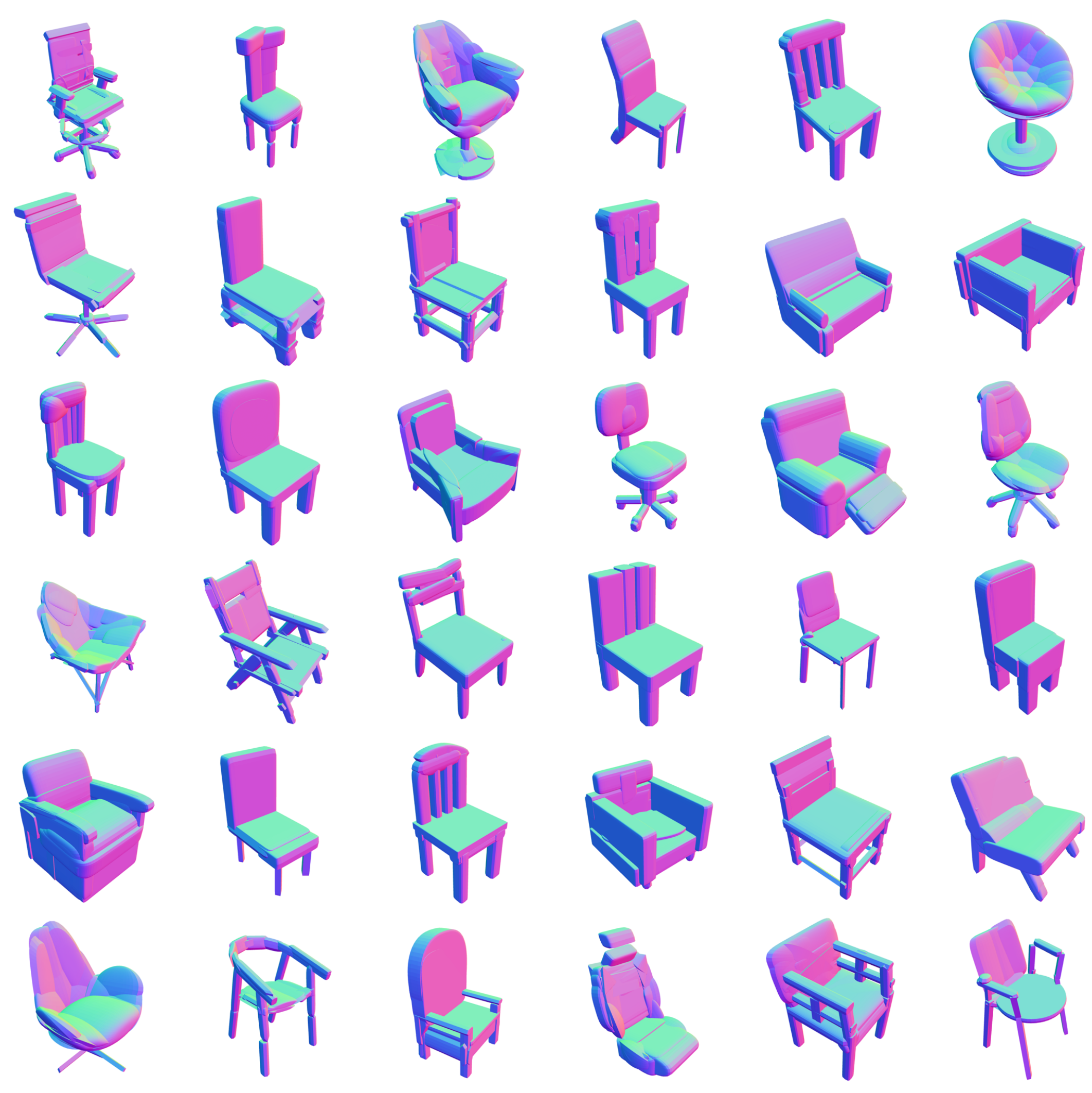}
  \captionof{figure}{Qualitative visualizations of high-fidelity and diverse results on \textsc{Chair} shape generation.}
  \label{fig:4}
\end{figure*}

\begin{figure*}
\centering
  \includegraphics[width=0.9\textwidth]{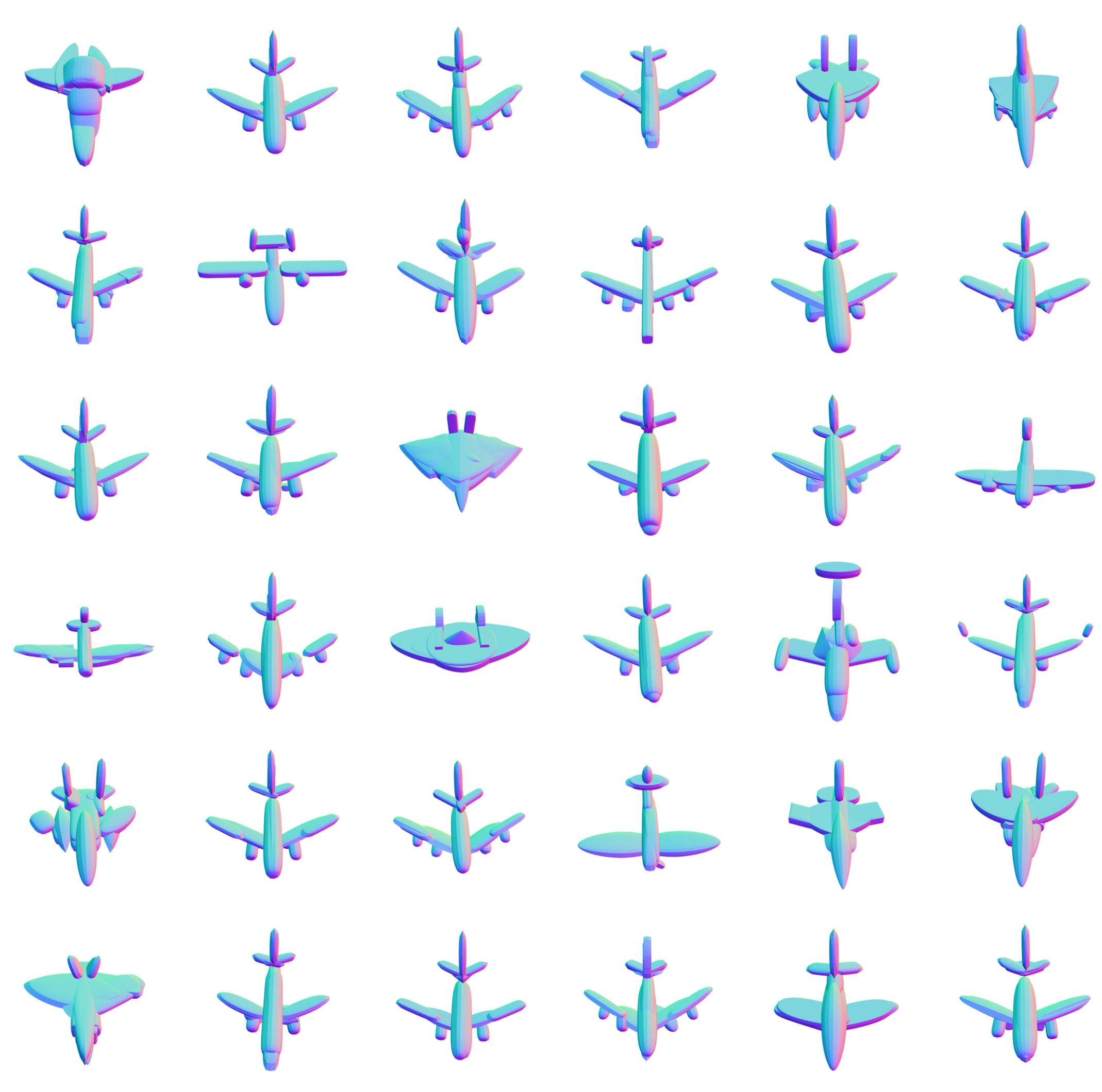}
  \captionof{figure}{Qualitative visualizations of high-fidelity and diverse results on \textsc{Airplane} shape generation.}
  \label{fig:5}
\end{figure*}

\begin{figure*}
\centering
  \includegraphics[width=0.9\textwidth]{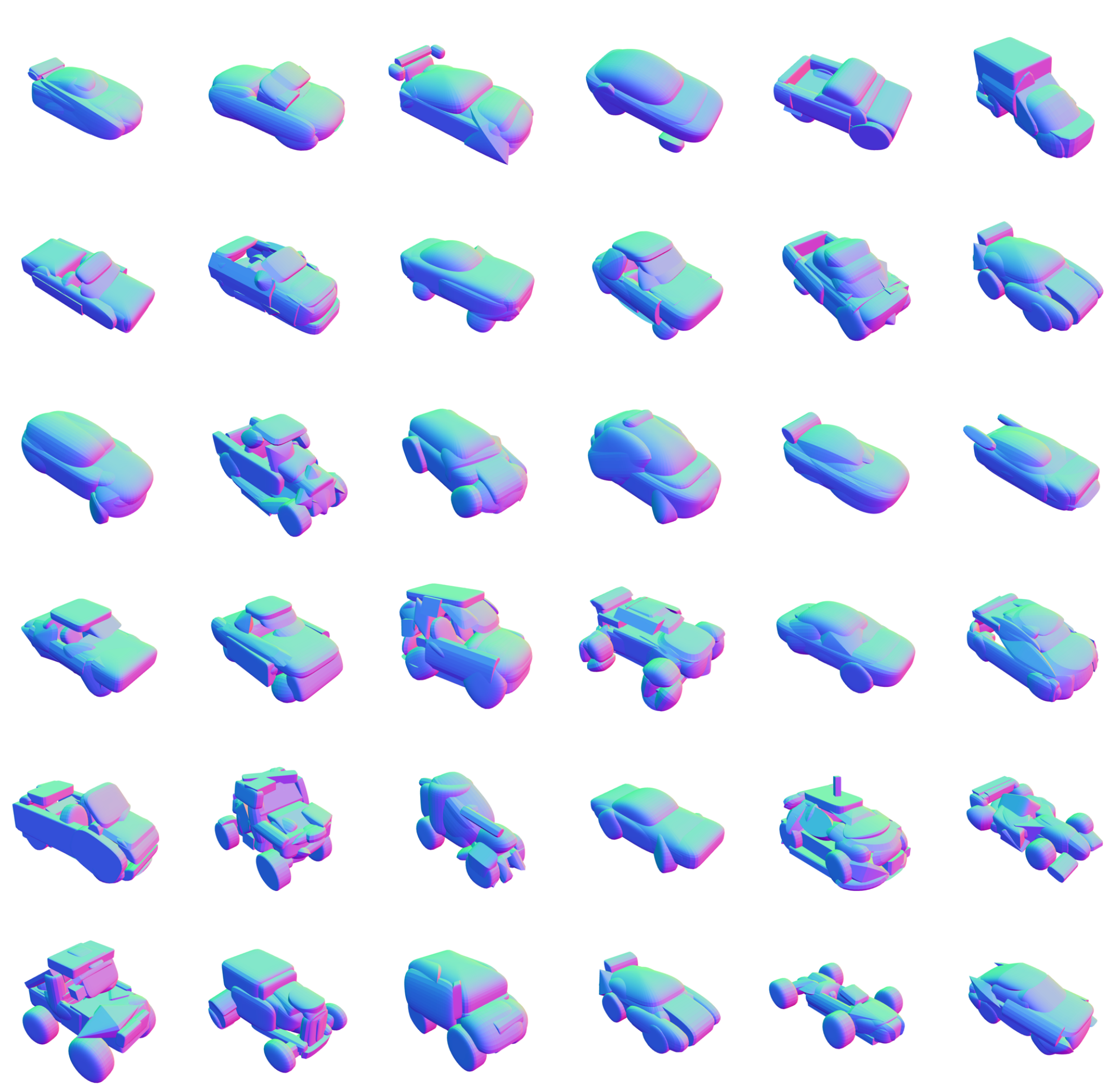}
  \captionof{figure}{Qualitative visualizations of high-fidelity and diverse results on \textsc{Car} shape generation.}
  \label{fig:6}
\end{figure*}

\end{document}